\documentclass[acmlarge]{acmart}
\usepackage{multirow}
\usepackage{algorithm}
\usepackage{algpseudocode}
\usepackage{placeins}
\usepackage{amsmath}

\usepackage{graphicx}
\usepackage{subcaption}

\expandafter\def\expandafter\normalsize\expandafter{%
  \normalsize  
  \setlength\abovedisplayskip{4ex}
  \setlength\belowdisplayskip{4ex}
  \setlength\abovedisplayshortskip{4ex}
  \setlength\belowdisplayshortskip{4ex}
}

\usepackage{etoolbox}
\BeforeBeginEnvironment{equation}{\noindent}

\pdfoutput=1

\AtBeginDocument{%
  \providecommand\BibTeX{{%
    \normalfont B\kern-0.5em{\scshape i\kern-0.25em b}\kern-0.8em\TeX}}}

\acmJournal{IMWUT}
\acmVolume{0}
\acmNumber{0}
\acmArticle{0 }
\acmMonth{0}

\begin{document}

\title{VCHAR:Variance-Driven Complex Human Activity Recognition framework with Generative Representation}

\footnote{This study was
approved by the Institutional Review Board at our institution (IRB
No. \phantom{Pro2018001757})}

  \author{YUAN SUN}
\email{s820@soe.rutgers.edu}
\affiliation{%
  \institution{WINLAB,Rutgers University}
  \city{Piscataway}
  \state{New Jersey}
  \country{USA}
}

  \author{NAVID SALAMI PARGOO }
\email{navid.salamipargoo@rutgers.edu}
\affiliation{%
  \institution{Rutgers University}
  \city{Piscataway}
  \state{New Jersey}
  \country{USA}
}

  \author{TAQIYA EHSAN}
\email{te137@rutgers.edu}
\affiliation{%
  \institution{Rutgers University}
  \city{Piscataway}
  \state{New Jersey}
  \country{USA}
}

  \author{ZHAO ZHANG}
\email{zz671@soe.rutgers.edu}
\affiliation{%
  \institution{Rutgers University}
  \city{Piscataway}
  \state{New Jersey}
  \country{USA}
}

  \author{JORGE ORTIZ}
\email{jorge.ortiz@rutgers.edu.}
\affiliation{%
  \institution{Rutgers University}
  \city{Piscataway}
  \state{New Jersey}
  \country{USA}
}

\renewcommand{\shortauthors}{ }

\begin{abstract}

Complex human activity recognition (CHAR) remains a pivotal challenge within ubiquitous computing, especially in the context of smart environments. Existing studies typically require meticulous labeling of both atomic and complex activities, a task that is labor-intensive and prone to errors due to the scarcity and inaccuracies of available datasets. Most prior research has focused on datasets that either precisely label atomic activities or, at minimum, their sequence—approaches that are often impractical in real-world settings. In response, we introduce VCHAR (Variance-Driven Complex Human Activity Recognition), a novel framework that treats the outputs of atomic activities as a distribution over specified intervals. Leveraging generative methodologies, VCHAR elucidates the reasoning behind complex activity classifications through video-based explanations, accessible to users without prior machine learning expertise. Our evaluation across three publicly available datasets demonstrates that VCHAR enhances the accuracy of complex activity recognition without necessitating precise temporal or sequential labeling of atomic activities. Furthermore, user studies confirm that VCHAR's explanations are more intelligible compared to existing methods, facilitating a broader understanding of complex activity recognition among non-experts.

\end{abstract}




\maketitle

\section{Introduction }
\label{chap:1}

% In recent years, the proliferation of sensors has become ubiquitous, seamlessly integrating into various aspects of our daily lives. These sensors are embedded in a myriad of devices and objects, ranging from smartphones and cameras to clothing, buildings, and vehicles. The continuous generation of massive amounts of data from these pervasive sensors has propelled the field of activity recognition to the forefront of ubiquitous computing research. The potential applications of activity recognition are vast, spanning domains such as healthcare\cite{do2013healthylife}, elderly care\cite{rashidi2012survey}, surveillance, and emergency response\cite{zhang2008assisting}, showcasing its immense utility and impact. However, this emerging field is not without its challenges. One of the primary obstacles\cite{zhao2011cross,pan2009survey,gomes2012mobile} lies in the labeling of sensor data, which often suffers from a lack of labels, incorrect labeling, or the need for manual annotation. Furthermore, many activity recognition models are often complex or opaque "black-boxes," hindering human interpretation, model transparency, and assessment\cite{atzmueller2010mining,ribeiro2016should,atzmueller2017onto}. However, AI experts' understanding doesn't guarantee laypeople's comprehension. Explainable AI methods tailored for non-technical individuals are crucial to clarify AI decision-making, increase transparency, and build trust. Accessible AI fosters confidence, facilitating wider adoption and integration in various domains\cite{zylowski2022study}.

In recent years, the proliferation of sensors across diverse settings has become ubiquitous, seamlessly integrating into the very fabric of daily life. These sensors are embedded in a wide array of devices such as smartphones, cameras, clothing, buildings, and vehicles, enabling continuous and pervasive data collection. This expansion has significantly propelled the field of activity recognition, placing it at the forefront of ubiquitous computing research. The potential applications of activity recognition are vast and impactful, encompassing areas such as healthcare \cite{do2013healthylife}, elderly care \cite{rashidi2012survey}, surveillance, and emergency response \cite{zhang2008assisting}. The ability to monitor and understand human activities on such a scale holds immense utility, offering advancements in personal health monitoring, enhanced security systems, and more efficient emergency services.

However, alongside its rapid growth and integration into various domains, the field of activity recognition faces several substantial challenges. Chief among these is the issue of sensor data labeling, which is crucial for the development of accurate and reliable models. This process often encounters significant hurdles such as the absence of labels, incorrect labeling, or the extensive manual effort required for annotation \cite{zhao2011cross,pan2009survey,gomes2012mobile}. Furthermore, the models developed for activity recognition frequently remain complex and opaque, functioning as "black-boxes" that are difficult for both experts and laypeople to interpret \cite{atzmueller2010mining,ribeiro2016should,atzmueller2017onto}. This complexity hinders transparency and assessment, posing a barrier to trust and understanding among users who are not well-versed in technical details.

To address these issues, there is a growing emphasis on the development of Explainable AI (XAI) methods. Such methods are crucial for demystifying AI decision-making processes, increasing transparency, and building trust among users \cite{zylowski2022study}. By making AI decisions more accessible and understandable, XAI not only enhances user confidence but also facilitates wider adoption and integration of these technologies across various sectors. This approach aims to ensure that as AI systems become more integrated into our lives, they do so in a manner that is both comprehensible and trustworthy, ultimately leading to more informed and accepting user interactions.

To unlock activity recognition's full potential in ubiquitous computing, addressing labeling challenges and creating techniques that enable layperson understanding of model insights is paramount. This bridges the gap between raw sensor data and actionable information, facilitating the development of reliable, trustworthy, and deployable real-world activity recognition systems.

\subsection{Challenges in Complex Human Activity Recognition}

Traditional methods for Complex Human Activity Recognition (CHAR) typically necessitate precise labeling of each atomic activity within specific time slots to effectively train models. While some research attempts to incorporate conceptual frameworks, these approaches often require segmenting the data to enhance accuracy. Such segmentation demands detailed labeling of each atomic activity, including the elimination of transient states, which can be labor-intensive and prone to inaccuracies regarding the exact start and end points of activities.

\begin{figure}[h]
  \centering
  \includegraphics[width=1\linewidth,,height=0.15\textheight]{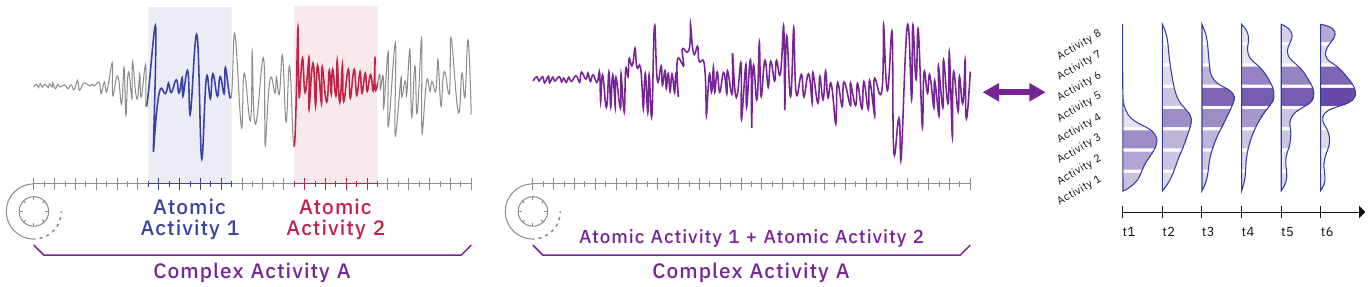}
\caption{Standard complex activity datasets (left) typically provide detailed labels for each time interval to facilitate atomic activity training. In contrast, in-the-wild datasets(middle), constrained by labor capacity and other practical limitations, only specify the types of complex and atomic activities per segment without specific time interval or detailed atomic activity labels for time series segmentation. It often feature a greater variety of label combinations(right), reflecting the complexity and unpredictability of real-world scenarios.}
  \label{labeling_issue}
\end{figure}

In practical scenarios, real-life datasets typically categorize types of atomic or complex activities within specific collection intervals (see Fig. \ref{labeling_issue})~\cite{hyzg-9m49-20,shoaib2016hierarchical,shoaib2016complex,chen2021deep}. While some datasets provide detailed atomic activity labels, these can often be erroneous or unreliable~\cite{inoue2019integrating,kwon2019handling}. Furthermore, some datasets only indicate the type of activities (Fig. \ref{labeling_issue}), encompassing \( n \) atomic activities where \( m \) activities may occur concurrently, leading to $C\mathbin{(}n,m\mathbin{)}$  possible combinations. This underscores the combinatorial complexity faced when segments cannot be distinctly separated. Our framework is specifically designed to address such challenges by managing inseparable dataset segments and extending beyond merely detecting \( n \) isolated activities. It is important to note that a prevalent assumption in the field suggests that the performance of machine learning models degrades as they become more explainable, especially when the model structures become intricate~\cite{gunning2016broad}. Additionally, these datasets often assign uncharacterized activities to generic categories like "others" or exclude them altogether, which presents significant labeling challenges and complicates the development of generalized solutions adaptable to real-world applications.

Moreover, significant challenges persist in representing the outputs of sensor-based models within the visual domain. Despite increasing interest in transforming sensor data into image representations to enhance layperson understanding of machine learning results, the development of visual domain representations has not kept pace\cite{arrotta2022dexar,hur2018iss2image,aquino2023explaining}. This discrepancy highlights the urgent need for innovative approaches that effectively bridge the gap between sensor data and visual representation, thereby improving the interpretability and practical utility of Complex Human Activity Recognition (CHAR) systems for everyday users. This growing interest emphasizes the demand for visual representations that make machine learning models more accessible to laypersons.

These challenges highlight the complexity of CHAR in real-life settings and underscore the necessity for advanced methodologies that can handle imprecise data and develop more intuitive output representations.

% \begin{figure}[h]
%   \centering
%   \includegraphics[width=1\linewidth,,height=0.2\textheight]{fig/simple_combination.pdf}
% \caption{Contrary to standard datasets, in-the-wild datasets often feature a greater variety of label combinations, reflecting the complexity and unpredictability of real-world scenarios.}

%   \label{labeling_issue_combination}
% \end{figure}

\subsection{Contributions}

We developed the Variance-Driven Complex Human Activity Recognition (VCHAR) framework with Generative Representation to tackle prevalent issues in the recognition of complex human activities. VCHAR overcomes the limitations of traditional CHAR methods that require detailed and labor-intensive segmentation of activities. Instead, it utilizes a variance-driven approach that leverages the Kullback-Leibler divergence to approximate the distribution of atomic activity outputs. This method allows for the recognition of decisive atomic activities within specific time intervals without necessitating the removal of transient states or other irrelevant data, thereby enhancing the detection rates of complex activities even in the absence of detailed labeling of atomic activities.Our experiments demonstrate that even without precise labeling of atomic activities or without sequentially corrected labeling, our model effectively utilizes key concepts to enhance the detection rate of complex activities. Additionally, it provides a promising rate of atomic activity detection, which is crucial for accurately representing the data when transmitting outputs to the decoder.

Moreover, VCHAR introduces a novel generative decoder framework that transforms sensor-based model outputs into integrated visual domain representations. This includes detailed visualizations of both complex and atomic activities, alongside desired sensor  related information from the model. Utilizing a Language Model (LM) agent, the framework organizes diverse data sources and employs a Vision-Language Model (VLM) to generate comprehensive visual outputs. To facilitate rapid adaptation to specific smart space scenarios, we propose a pretrained "sensor-based foundation model" and implement a "one-shot tuning strategy" with masked guidance. Our experiments on three publicly available datasets demonstrate that VCHAR not only performs competitively with traditional methods but also significantly enhances the interpretability and usability of CHAR systems, as confirmed through human evaluation studies.Our contributions are multi-faceted and can be summarized as follows:

% a frame work, 
% variance driven,kl,multi task wont compermise the detetion rate, helps to improve
% genertive decoder , adaptible to diverse senario
% evaluation

\begin{itemize}
    \item We introduce VCHAR, a variational-driven framework designed to generate visual domain representations of complex activity recognition. This system aims to make complex activity insights accessible to laypersons by visually representing the data in an intuitive manner.
    
    \item  We utilize KL divergence as a loss function to model the dynamic relationships among varying combinations of atomic activities across different time intervals for the same type of complex activity.

    \item  Our method proves effective in real-life scenarios with inaccurate or absent specific time labeling of activities. Results demonstrate that multitasking modeling enhances complex activity detection rates. Additionally, this multitasking modeling equips the decoder with features necessary to offer visual domain explanations accessible to laypersons.
    
    \item  We propose a "sensor-based foundation model" framework with our masked one-shot tuning strategy that quickly adapts to specific smart space scenarios. An LLM agent guides this model to generate accessible visual domain representations, particularly benefiting users without technical expertise.

     \item  We conducted experiments on 3 publicly available datasets, some with labeling issues. Our method demonstrated competitive results and user preference through a user study, showcasing its effectiveness.
    
\end{itemize}

% These contributions collectively advance the state of the art in CHAR by not only improving recognition accuracy but also by enhancing the interpretability, usability, and adaptability of recognition systems in ubiquitous computing environments.

\begin{figure}[h]
  \centering
  \includegraphics[width=1\linewidth,height=0.3\textheight]{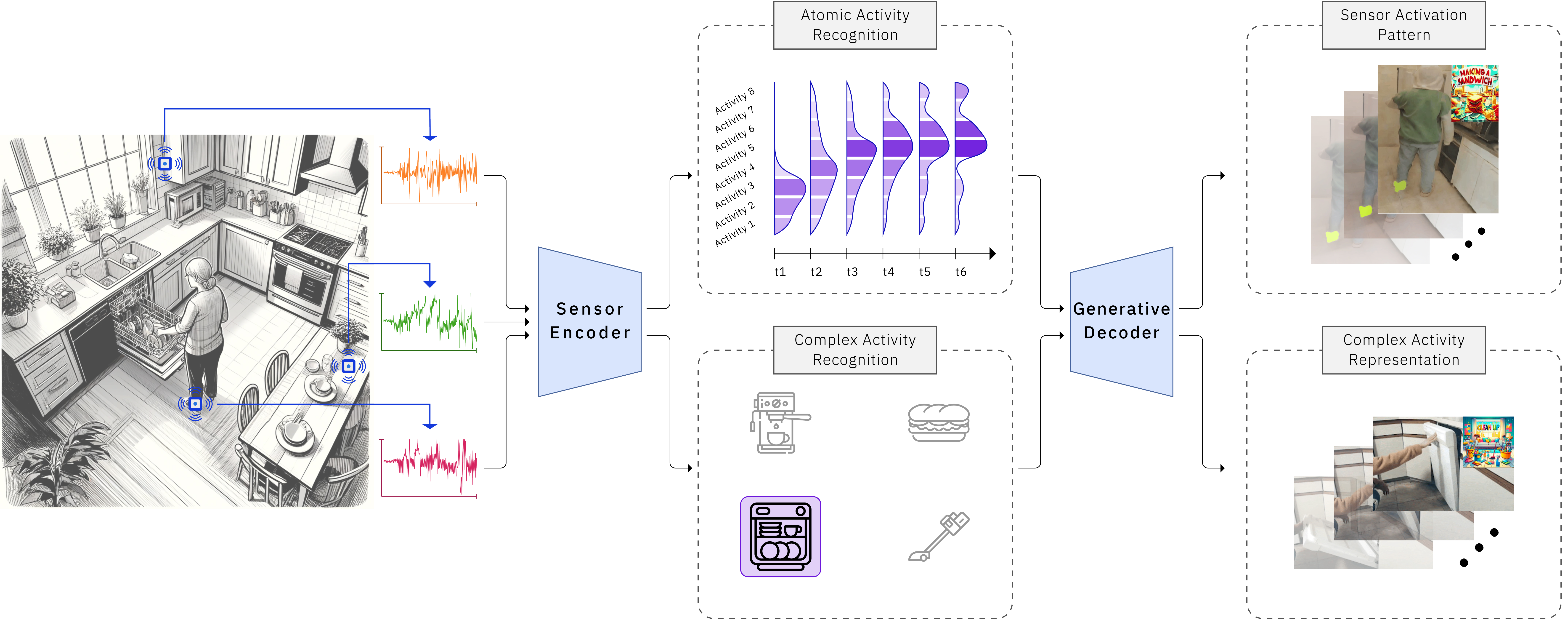}
 \caption{The VCHAR framework is tailored for recognizing both atomic and complex activities without precise temporal annotations. The framework leverages sensor encoder outputs to generate visual representations at 24 fps, thereby enhancing user understanding. For instance, one video output depicts how the left foot sensor crucially detects activities. Another segment illustrates the atomic activity ``open the dishwasher'' as part of the broader ``clean up'' complex activity.}
  \label{overall_result}
\end{figure}

\section{Related work}
\label{chap:2}

\subsubsection{Smart Space Complex Activity Recognition }

% Bao et al\cite{bao2004activity}. used multiple biaxial accelerometers and decision tree classifiers to recognize a variety of everyday activities, from atomic to complex, with high accuracy. They found that more accelerometers and subject-specific training improved performance.

% Dernbach et al\cite{dernbach2012simple}. used a triaxial accelerometer and gyroscope in a smartphone with a Multi-layer Perceptron classifier to recognize both atomic and complex activities, finding that atomic activities were accurately classified but complex activities posed challenges.

% Mekruksavanich et al\cite{mekruksavanich2021deep}. proposed a CNN-BiGRU model that combines convolutional neural networks (CNNs) and bidirectional gated recurrent units (BiGRUs) to recognize complex human activities from wrist-worn sensor data

% Tahvilian et al.\cite{tahvilian2022accuracy} compared CNN-LSTM and CNN-BiLSTM models for recognizing atomic and complex human activities. 

% Peng et al. proposed \cite{peng2018aroma}, a deep multi-task learning approach using CNNs and LSTMs to jointly recognize atomic and complex activities from accelerometer, gyroscope, and magnetometer data. They found that sharing a CNN feature extractor between the tasks improved performance on both compared to single-task models.

%  complex activities, being more intricate, are harder to recognize than atomic ones, resulting in lower performance.\cite{peng2018complex}.

Significant advancements have been made in recognizing both atomic and complex human activities using various sensor technologies and machine learning models. Bao et al. utilized multiple biaxial accelerometers along with decision tree classifiers, demonstrating that an increase in the number of accelerometers and subject-specific training enhances performance \cite{bao2004activity}. Dernbach et al. found that while atomic activities could be accurately classified using a smartphone's triaxial accelerometer and gyroscope with a Multi-layer Perceptron, complex activities posed greater challenges \cite{dernbach2012simple}. Mekruksavanich et al. introduced a CNN-BiGRU model that effectively recognizes complex activities from wrist-worn sensors \cite{mekruksavanich2021deep}, while Tahvilian et al. compared the efficacy of CNN-LSTM and CNN-BiLSTM models for varying complexities of activities \cite{tahvilian2022accuracy}. Additionally, Peng et al. proposed a multi-task learning approach using CNNs and LSTMs that improved performance on both atomic and complex activities by sharing a CNN feature extractor \cite{peng2018aroma}. They also noted that complex activities, due to their intricate nature, are harder to recognize than atomic ones, resulting in lower performance \cite{peng2018complex}. This body of work collectively emphasizes the potential of integrating advanced computational models and diverse sensor data to enhance activity recognition in ubiquitous computing environments.

\subsubsection{Visual Representation of Sensor Data}

The transformation of sensor data into visual formats has gained significant attention, enabling the application of image classification techniques to sensor-based human activity recognition \cite{ha2016convolutional, trabelsi2022sensor}. Researchers have explored various methods to facilitate this transformation. For instance, transforming sensor data into spectrogram images has allowed the use of deep learning models like CNNs for recognizing human activities \cite{ito2018application,jiang2015human,chen2021deep}.

Further advancements in this field have led to the development of methods that enhance interpretability. Arrotta et al.\ \cite{arrotta2022dexar} utilize CNNs and Explainable AI to translate sensor data into semantic maps for transparent activity recognition in smart homes, aiming to boost trust, particularly in healthcare monitoring scenarios. These semantic maps are further processed into natural language explanations, providing clear and understandable insights into the AI's decision-making processes. Another approach by Jeya et al.\ \cite{jeyakumar2023x} applies CTC loss to align detected activities with raw sensor signals, which are visualized on time-acceleration graphs and marked with dashed rectangles to improve the interpretability of complex activities.

Additionally, explainability in AI can be approached through model-agnostic methods such as LIME \cite{chattopadhay2018grad} and SHAP \cite{lundberg2017unified}, which approximate the relationship between input and output predictions without accessing the model's internal workings. Conversely, model-transparent methods like Grad-CAM++ \cite{chattopadhay2018grad} and saliency maps \cite{simonyan2013deep} provide insights into the internal processes of neural networks by visualizing the importance of input features and the activation value of hidden layers.

In response to the growing interest in making sensor data comprehensible through visual representations\cite{hur2018iss2image}, our work develops vision domain representations directly from sensor data. Specifically, we aim to make the outcomes of activity recognition models accessible to laypersons by applying a video-based representation of sensor activation value. This approach bridges the gap in understanding complex models for users without technical expertise, enhancing user trust and engagement with the technology.
\subsubsection{Foundation and Multimodal Models}

Foundation models, the latest evolution in AI technologies, are trained on extensive, diverse datasets and are capable of being applied across a wide range of domains \cite{bommasani2021opportunities}. These models, which are adaptable to numerous applications, highlight the forefront of AI research, benefiting from their training on vast and heterogeneous datasets.

Recent advancements have seen various multimodal models that leverage this foundation. Wang et al. \cite{wang2022ofa} developed a  sequence-to-sequence framework that unifies diverse tasks across modalities using an instruction-based task representation, pretrained on image-text data for both crossmodal and unimodal tasks. Similarly, Lu et al.\ \cite{lu2022unified} introduced a transformer sequence-to-sequence model that performs a variety of vision and language tasks without requiring task-specific branches, trained on over 90 datasets related to vision and language.

Furthering the multimodal approach, Singh et al.\ \cite{singh2022flava} created a model with separate encoders for images, text, and multimodal integration, which was pretrained on unimodal and multimodal losses for 35 tasks across vision, language, and vision-language areas. Another contribution by Singh et al.\ \cite{wang2022image} includes a multimodal vision-language model with a shared multiway transformer backbone, trained on masked data modeling across modalities, achieving state-of-the-art performance on various vision and vision-language benchmarks. Li et al.\ \cite{li2020unimo} explored cross-modal contrastive learning to unify representations across modalities with a unified transformer on image and text data, enhancing the synergy between these modalities.

In this research, we harness the domain adaptation capabilities of generative models to customize the sensor decoder for specific scenarios of sensor data representation. This approach is designed to significantly enhance the visualization quality of the sensor model's outputs.

\section{Research Methods}
\label{chap:3}

This section systematically details the VCHAR framework, commencing with an architectural overview and operational functionality. We begin by outlining the fundamental architecture and principal features of VCHAR, providing a foundation for detailed exploration. This is followed by an in-depth analysis of the conceptual framework that forms the basis of VCHAR, discussing both the formulation of the problem it addresses and the integration of relevant theoretical concepts. Through this structured exposition, we aim to furnish a comprehensive understanding of VCHAR’s underlying principles and its functionality within practical applications.

\subsection{Outline of the VCHAR Framework}

\begin{figure}[h]
  \centering
  \includegraphics[width=1\linewidth,,height=0.3\textheight]{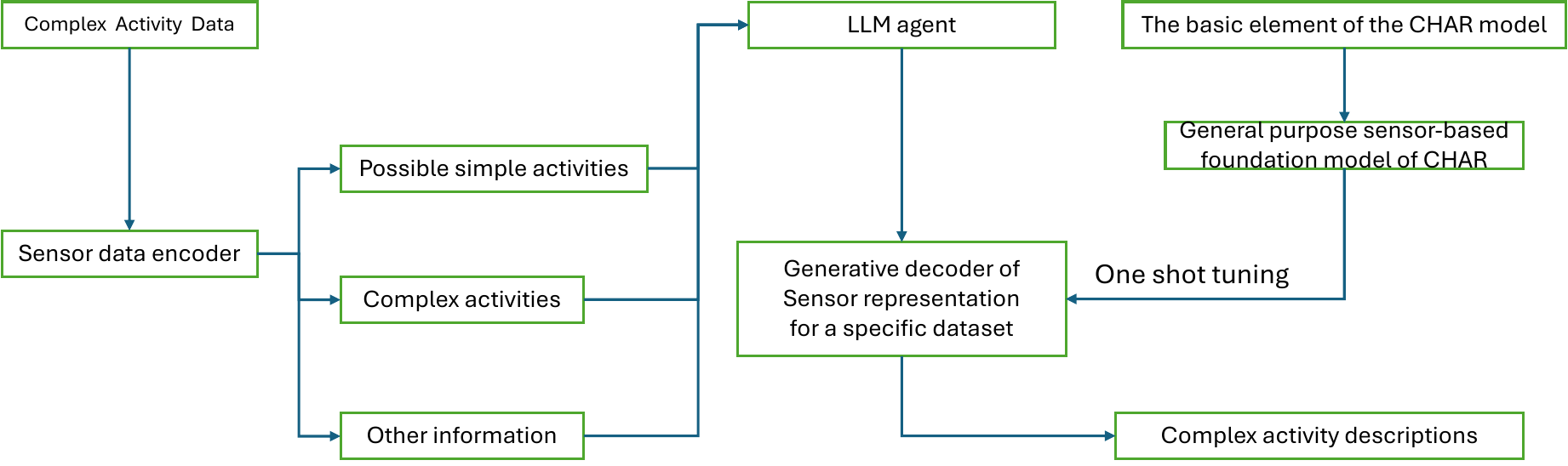}
 \caption{The VCHAR framework integrates various types of sensor data and utilizes a LLM agent to structure this information for a generative decoder. The framework is designed to enhance user comprehension by providing additional insights such as sensor channel activation values, offering a deeper understanding of the events occurring within the smart space.}

  \label{overall_structure}
\end{figure}

The Variance-Driven Complex Human Activity Recognition (VCHAR) framework, as depicted in Fig.~\ref{overall_structure}, is an end-to-end model designed to enhance both the prediction and explanation of complex human activities. This model uniquely employs the Kullback-Leibler (KL) divergence to approximate the distribution of atomic activities across various sliding window lengths. By leveraging a variance-driven approach, VCHAR circumvents the need for specific time-unit labels for atomic activities, focusing instead on the types of activities occurring within given time slots. Comparative experiments demonstrate that this approach significantly enhances the accuracy of complex activity detection relative to other methods.

The multitask design of VCHAR not only improves the recognition of complex activities but also establishes an "interface" facilitating detailed visualizations by a generative decoder. For instance, when VCHAR detects complex activities such as making a sandwich, it concurrently identifies related atomic activities like opening a door or turning a switch. It also provides a list of desired sensor related information from the model, highlighting the significance of each sensor in the smart space environment. 

To further enrich the model's output, a Language Model (LM) agent is integrated to reorganize and elucidate detailed information about these activities. In practical applications, we propose a "sensor-based smart space foundation model" framework, capable of being tailored to specific scenarios through advanced techniques. Techniques such as Denoising Diffusion Implicit Models (DDIM) and Latent Diffusion Models (LDMs), coupled with a masked training strategy, are employed to refine the quality and relevance of the generated outputs.

This structured process ensures that VCHAR is not only effective in recognizing complex human activities but also adept at providing actionable insights into the dynamics of smart spaces, thereby enhancing user interaction and understanding.

\subsection{Multi-Task Learning for Complex Activity Recognition}

In the proposed model, multi-task learning is employed to facilitate the simultaneous recognition of atomic and complex activities. Atomic activities are defined as discrete actions that occur within a brief time window and are indivisible in the context of our dataset. Each complex activity, on the other hand, is composed of multiple atomic activities. Specifically, a complex activity in our model is defined as having more than 2 atomic activities, providing a more nuanced description of the grouped atomic activities.

For instance, within the Opportunity dataset \ref{opportunity_dataset}, complex activities such as "making coffee" or "cleaning up" may include the atomic activity "open the door". This activity is further detailed as "open the door 1" and "open the door 2", enriching the model's understanding of the scenario by supplying detailed contextual information. This approach allows the model to capture the intricate relationships and recurring patterns among activities, thereby enhancing its ability to accurately classify complex activity scenarios.

Our model analyzes raw sensor data, denoted by \( \mathbf{x} \), to predict the probabilities of each atomic activity occurring within a specific sliding window, alongside the classification of a complex activity. The goal is to output a probability vector \( \mathbf{p} \) for atomic activities, where each element \( p_i \) represents the likelihood of the \( i^{th} \) atomic activity occurring. Additionally, the model outputs a categorical label \( C \) that classifies the type of complex activity observed, which integrates the information from the atomic activities.

Formally, \( \mathbf{x} \) represents the input vector of sensor readings. The output \( \mathbf{p} \) is a vector of probabilities, with length \( n \), where \( n \) is the total number of atomic activities the model is trained to recognize. Each element \( p_i \) in \( \mathbf{p} \) is a real number between 0 and 1, inclusive, indicating the probability that the \( i^{th} \) atomic activity is present in the sliding window. The complex activity label \( C \), determined by these probabilities, provides a high-level classification based on the pattern of atomic activities.

For instance, if the model considers four atomic activities, and a particular observation through the sliding window suggests varying probabilities of these activities, the output vector \( \mathbf{p} \) might look like \([0.95, 0.80, 0.10, 0.05]\). This vector indicates high probabilities for the first two activities and low probabilities for the others. The label \( C \) then contextualizes these activities into a complex activity classification, providing a comprehensive understanding of the scenario captured by \( \mathbf{x} \).

The primary training objective of our model is to minimize a loss function, \(L\), which effectively combines the losses from predicting the probabilities of atomic activities, \(L_{atomic}\), and from classifying complex activities, \(L_{complex}\). This combined loss function is defined as:

\[
L = \alpha L_{atomic}(p_{atomic}, p_{true}) + \beta L_{complex}(y_{complex}, C_{true})
\]

In this formula, \(\alpha\) and \(\beta\) are weighting coefficients that balance the importance of each component during the training process. Here, \(p_{true}\) represents the true distribution of the atomic activities occurring within a given context, and \(C_{true}\) is the actual label of the complex activity. The dual-focus of this loss function encapsulates the essence of our multi-task learning approach, promoting an efficient and robust learning process that is well-suited for analyzing the nuanced dynamics of smart space sensor data. This methodology not only improves the predictive accuracy of both atomic and complex activity classifications but also ensures that the model can effectively discern the intricate relationships between these activity layers.

This modeling approach, with its probabilistic output for atomic activities, primarily facilitates the detection and understanding of complex activities, the main objective of our model. By analyzing the likelihoods of various atomic activities within a given context, our model uses these insights as supportive data to enhance the accuracy and reliability of complex activity classification. This method allows for a more nuanced interpretation of sensor data, ensuring that atomic activities serve to inform and refine our understanding of the broader, more intricate behavioral patterns represented by complex activities.

\subsection{Loss of Atomic Activity Recognition }

% Our model is tasked with predicting the likelihood of each atomic activity within a given sliding window of sensor data. The aim is to ensure that the model's predictions closely match the actual occurrence probabilities using the Kullback-Leibler (KL) divergence as the loss function. KL divergence quantifies the difference between the predicted probability distribution \( p_{\text{predict}} \) and the true probability distribution \( p_{\text{true}} \), serving as an effective metric for evaluating prediction accuracy in terms of information loss.

% Formally, the loss function \( L \) for the atomic activities is defined using the KL divergence between the true distribution \( p_{\text{true}} \) and the predicted distribution \( p_{\text{predict}} \), expressed as follows:

% \[
% L_{KL}(p_{\text{predict}}, p_{\text{true}}) = \sum_{i} p_{\text{true}, i} \log \frac{p_{\text{true}, i}}{p_{\text{predict}, i}}
% \]

% Here, \( p_{\text{true}, i} \) and \( p_{\text{predict}, i} \) denote the true and predicted probabilities of the \( i^{th} \) atomic activity occurring within the sliding window, respectively. Minimizing this loss during training optimizes the model's ability to accurately predict the probabilities of atomic activities.

% In parallel, the model classifies complex activities based on these probability predictions, thereby enriching our understanding of the sensor data's nuanced dynamics. This integrated approach not only sharpens the model’s activity recognition capabilities but also boosts its overall interpretability and efficacy in complex environmental settings.

Our model's primary objective is to predict the probability of each atomic activity within a sliding window of sensor data, aligning these predictions as closely as possible with the actual probabilities using the mean Kullback-Leibler (KL) divergence as the loss function. The  KL divergence provides a robust metric for the average difference between the predicted probability distribution \( p_{\text{predict}} \) and the true probability distribution \( p_{\text{true}} \), making it particularly suitable for datasets with varying class distributions.

Formally, the loss function \( L \) for atomic activities is defined as the  KL divergence between the true distribution \( p_{\text{true}} \) and the predicted distribution \( p_{\text{predict}} \), which can be expressed mathematically as:

\[
L_{atomic}=L_{KL}(p_{\text{predict}}, p_{\text{true}}) = \frac{1}{N} \sum_{i=1}^{N} p_{\text{true}, i} \log \frac{p_{\text{true}, i}}{p_{\text{predict}, i}}
\]

where \( N \) is the number of classes or atomic activities, \( p_{\text{true}, i} \) and \( p_{\text{predict}, i} \) represent the true and predicted probabilities of the \( i^{th} \) atomic activity occurring within the sliding window, respectively. This formulation ensures that the model's performance is evaluated based on the average divergence across all classes, promoting a balanced sensitivity to the accuracy of each class prediction.

\subsection{ Loss of Complex Activity Recognition}

Our model is specifically designed to classify complex activities by minimizing the cross-entropy loss, which measures the discrepancy between the predicted probabilities and the actual class labels for complex activities. This loss function is crucial for optimizing the model's ability to accurately categorize complex activities based on sensor data inputs.

The cross-entropy loss for complex activity classification is formally defined as:

\[
L_{\text{complex}}(y_{\text{predict}}, C_{\text{true}}) = -\sum_{j=1}^{M} C_{\text{true}, j} \log y_{\text{predict}, j}
\]

where,\( y_{\text{predict}} \) represents the predicted probability distribution across the complex activity classes.\( C_{\text{true}} \) is the one-hot encoded vector of the true class labels for the complex activities.\( M \) is the number of possible complex activity classes. Minimizing \( L_{\text{complex}} \) during training ensures that the predicted probabilities align closely with the true class labels, effectively enhancing the model’s accuracy in complex activity recognition.

\subsection{Sensor Encoder Architecture}
\label{sub_section_encoder}

\begin{figure}[h]
  \centering
  \includegraphics[width=0.8\linewidth,,height=0.2\textheight]{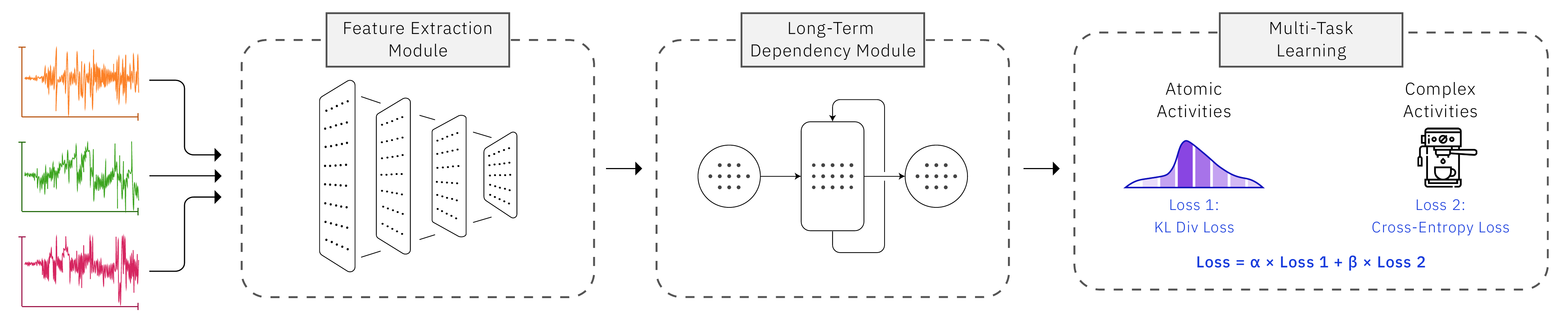}
\caption{For comparative analysis, the sensor encoder structure predominantly utilizes the ConvLSTM module. This encoder is designed to identify possible atomic activities and their associated complex activities occurring within the smart space.}
  \label{encoder_structure}
\end{figure}

To benchmark our approach against established baseline methodologies\cite{jeyakumar2023x,chen2021deep,peng2018aroma,singh2020deep}, we employ the widely recognized ConvLSTM architecture, which is particularly adept at handling sensor time series data. This architecture synergistically combines Convolutional Neural Networks (CNNs) and Long Short-Term Memory (LSTM) networks to effectively extract and temporally analyze features from sensor data.

The ConvLSTM architecture operates in two primary phases:
\begin{enumerate}
    \item \textbf{Feature Extraction:} The CNN component is responsible for spatial feature extraction from each time slice of the sensor data. This step is crucial for identifying intricate patterns within the data that are spatially localized but temporally variant.
    \item \textbf{Temporal Dependency Modeling:} The LSTM layer processes the sequence of extracted features to capture temporal dependencies and dynamics, essential for understanding the progression and context of sensor readings over time.
\end{enumerate}

In our enhanced model, we first conduct a channel-wise analysis to separately study the features from different sensor channels. These features are then integrated using a sensor fusion module, which synthesizes information across channels to provide a comprehensive feature set.

Following the fusion step, a bidirectional LSTM (biLSTM) module is employed to further refine the temporal analysis, enhancing the model's ability to capture both forward and backward dependencies in the time series data. Additionally, a Multi-Layer Perceptron (MLP) module is incorporated to generate the distribution of atomic activities based on the extracted features. Finally, another LSTM layer is tailored to model and predict the complex activity outputs, synthesizing all prior analyses into a coherent activity prediction.

\textbf{Example Application:}
Consider a scenario involving the monitoring of elderly activities in a smart home environment. Our model processes data from various sensors (e.g., motion, door, and appliance usage sensors) through the described architecture. Initially, individual sensor channels are analyzed to detect basic movements and interactions. These are then fused and temporally analyzed to predict more complex activities, such as cooking or cleaning, demonstrating the model’s capability to discern nuanced human behaviors effectively.

This comprehensive approach allows us to not only match but also surpass the performance of existing methods in complex activity recognition, as evidenced by our comparative evaluations. The results confirm the superiority of our model in accurately detecting and predicting both atomic and complex activities, highlighting its potential for real-world application in ubiquitous computing environments.

\subsection{Generative Modeling for Enhanced Complex Activity Representation}

\begin{figure}[h]
  \centering
  \includegraphics[width=\linewidth,,height=0.3\textheight]{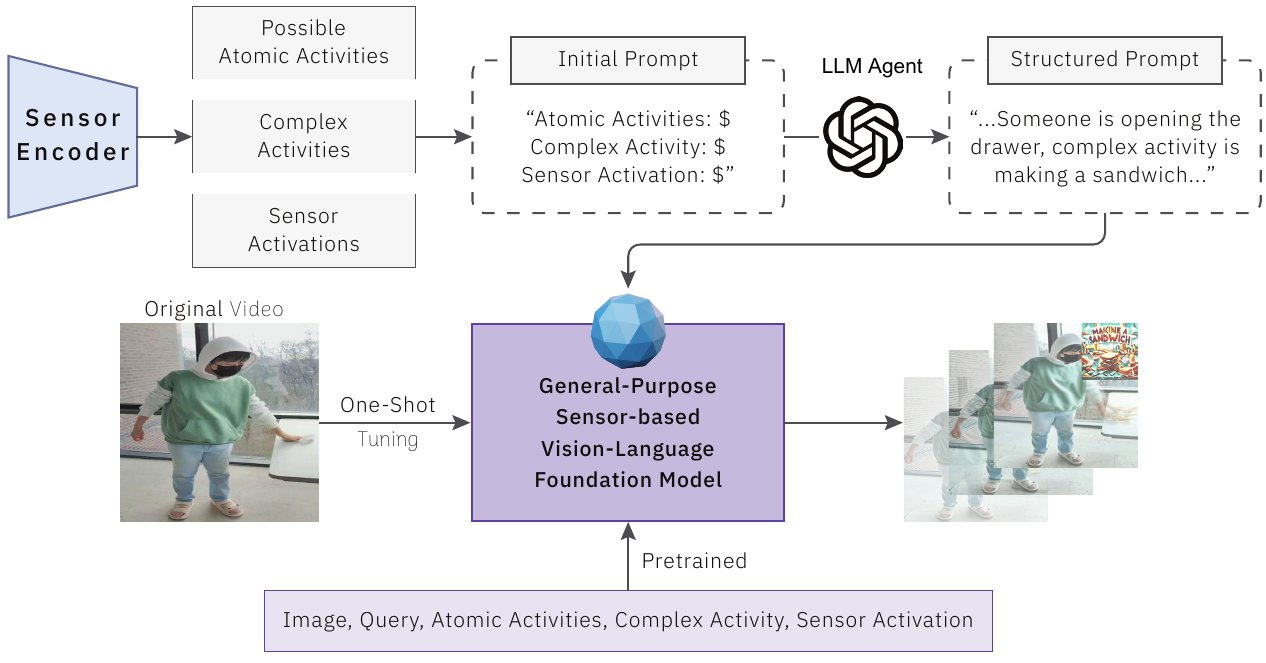}
\caption{The complex activity prompt learning decoder is engineered to master key elements including scenario descriptions, concept relationships, and detailed activity insights. This decoder is designed for adaptability, employing a one-shot tuning strategy to seamlessly integrate with specific datasets, thereby enhancing its versatility across various settings. The example shown in the graphs illustrates how this framework can be applied to newly identified activities, such as "clean the table," allowing the decoder to generate an intricate description of "making sandwich" tailored to the new dataset.}

  \label{diffusion}
\end{figure}

For users lacking technical expertise, grasping the intricacies of sensor encoder outputs can be challenging. To bridge this gap, we implement a generative modeling approach that transforms the identified atomic and complex activities into visual narratives. This transformation is facilitated by a Language Model (LM) agent, which interprets the sensor data, encompassing the distribution of atomic activities, the classification of complex activities, and the sensor activation patterns within the model.

To enhance the adaptability of our system across various datasets and smart spaces, we initially pre-train our model on a diverse set of scenarios. This pretraining encapsulates a universal concept of relationships and fundamental elements essential for activity recognition. When adapting to a specific dataset or smart environment, we employ the "one-shot tuning strategy". This approach fine-tunes the pre-trained model, enabling it to generate tailored explanations that align with the unique context and requirements of the given application.

%\subsubsection{ sensor to video generation pre-train :  Prompt learning with foundation models }
% \subsubsection{ Pre-train a general purpose sensor based foundation model of CHAR }

% As most of the foundation models, we firstly want to train a sensor based foundation model that contains basic elements for most of the senario. And then , when the user use it in a specific senario, it can just apply finetune technique to adapt the model to a specific scenario. The basic element for sensor based foundation model of CHAR should include most of the description of the complex senario, human activity. The different part is the model should also include the specific pattern of compex activity description and for model understanding purpose, specific sensor that is important to the model detection should also be visualized in the output. 

% for example, in our case, complex activity description like "morning rountine", "make a sandwich","making a cereal", can be pretrained in the foundation model. The description pattern can be a complex description in a certain area of the atomic activity description. The model should also be able to understand the sensors in the smart space that contribute most when the model detect the activity. These kinds of information can provide abundant details for the users without expertise

\subsubsection{Pretraining a General-Purpose Sensor-Based Foundation Model Framework for CHAR}

In line with the development of robust foundation models, our approach involves pretraining a sensor-based foundation model encapsulating a comprehensive suite of elements common to a wide array of scenarios. This model serves as a versatile starting point, designed to be fine-tuned subsequently to accommodate the specificities of distinct scenarios encountered by users.

The CHAR foundation model is imbued with a rich lexicon that describes a multitude of complex scenarios and human activities. It not only captures the essence of activity patterns but also identifies sensor-specific signatures that are pivotal for accurate activity detection. This capability ensures that, upon detection, the most relevant sensors are highlighted, providing intuitive visual cues within the generated explanations.

For instance, the foundation model is pretrained with elaborate activity narratives such as "morning routine", "making a sandwich", or "preparing cereal". These narratives embody complex sequences situated within a broader context of atomic activity representations. Moreover, the model is attuned to discern and emphasize sensors that yield significant information during the activity recognition process. This level of detail furnishes users lacking domain expertise with rich, contextual insights into the sensor data and the detected activities.

The foundation model thus constitutes a preparatory step in building an adaptable, knowledgeable system for CHAR. Through fine-tuning, it can swiftly assimilate into any designated smart space environment, delivering bespoke explanations tailored to the unique configuration and operation of that space.

% the objective function in the pretraining phase can be described as minmize the expectation of loss between ground truth and the output of the foundation model.the input of the foundation model is the image or vide, text, output from sensor encoder.

During the pretraining phase, the foundation model's objective function aims to minimize the expected discrepancy between the model's predictions and the ground truth. The foundation model processes three distinct inputs: the visual content \( V \), represented by images or videos; the query \( Q \), which contextualizes the visual content; and the sensor encoder output \( E \), which encapsulates the sensor-derived information. The objective function is thus characterized by the following minimization:

\[
\min \mathbb{E}_{V, Q, E \sim \mathcal{D}} \left[ \mathcal{L}\left(\text{GT}, \Psi_{\theta}\left(V, Q, E\right)\right) \right]
\]

Here, \( \Psi_{\theta} \) denotes the predictive function of the foundation model parameterized by \( \theta \), \( \text{GT} \) represents the ground truth corresponding to the inputs, and \( \mathcal{D} \) signifies the joint distribution of the visual, query, and sensor data. This optimization ensures that the foundation model is capable of integrating and interpreting multimodal inputs to generate a comprehensive and accurate depiction of the observed activities in preparation for fine-tuning to specific smart environment applications.

In our approach, \( E \) symbolizes a set of features including the type of activities identified, the values of sensor activation, among others. To enhance interpretability, we convert these features into a structured prompt using a predefined template. For instance, given an input vector \( [0.1, 0.2, 0.8, 0.5] \) where the third value indicates the highest sensor activation value from the model, and knowing that the corresponding sensor is located on the arm, our template highlights this sensor in yellow within the generated video. Concurrently, the model identifies the complex activity, such as 'making coffee', and a atomic activity, such as 'opening a door'. These elements are woven into a cohesive prompt: "Someone is opening the door ,complex activity "making coffee" ". This narrative integration allows for a user-friendly representation of activities, facilitating a clear and intuitive understanding of the ongoing events as detected by the sensor system.

\subsubsection{One-Shot Fine-Tuning for Complex Activity Description Using DDIM }

When transitioning our model to specific scenarios, it is common to encounter variations in the manifestation of activities and their corresponding complex contexts. To accommodate these scenario-specific nuances, we adopt a "one-shot tuning" strategy\cite{wu2023tune}. This strategy rapidly recalibrates our sensor-based foundation model to align with the new scenario characteristics. This is particularly pertinent for incorporating new activity videos where the contextual dynamics may significantly differ.

To further enhance the model's adaptability and descriptive capabilities, we introduce a masked training strategy. This approach facilitates the model's ability to generalize across diverse descriptive modalities. During fine-tuning, carefully designed prompts address the functionalities of specifically masked regions within the video. These prompts serve a dual purpose: they guide the video generation process in the latent space and ensure that the model's output is both representative and specific to the newly adapted scenario.

Employing this method enables the foundation model to not only recognize and adapt to new scenarios but also to generate activity representations that are rich in detail and contextually relevant.

Latent diffusion models (LDMs)\cite{rombach2022high} are pivotal in our approach, functioning directly within the latent space of new video embeddings. These models utilize an encoder to project videos into a latent space, facilitating manipulations within this reduced dimensionality before reconstruction. Specifically, an encoder \( \mathcal{E} \) maps a video \( V \) onto a latent representation \( \mathbf{z} = \mathcal{E}(V) \), and a decoder \( \mathcal{D} \) subsequently restores this latent representation back to the video domain, approximating the target video, i.e., \( \mathcal{D}(\mathcal{E}(V)) \approx V \).

In our case, when new activity types are introduced to the model, the LDMs adeptly handle the designated masked regions within the latent embeddings. The decoder \( \mathcal{D} \) is then employed to regenerate the video, ensuring that it accurately reflects the complex activity type and sensor activation  value in side of the model.

\[
\min_{\mathcal{E},\mathcal{D}} \left\| V - \mathcal{D}(\mathcal{E}(V)) \right\|^2
\]

DDIM, the Denoising Diffusion Implicit Model, is integral to enhancing the efficiency of the video generation process, particularly in maintaining the structural integrity of the activity movements\cite{wu2023tune,song2020denoising}.In our implementation, the DDIM operates within the latent space of a new video's embedding. The process utilizes prompts that encapsulate masked patterns, denoted by \( T \), alongside a guiding prompt \( T^* \) for directed generation. The final video \( \mathbf{V} \) is produced by first inverting the latent embedding via \( \text{DDIM}_{\text{inv}} \), using the encoder's output informed by the mask pattern and guiding prompt, followed by sampling with \( \text{DDIM}_{\text{samp}} \). This sequence can be represented mathematically as:

\[
\mathbf{V} = \mathcal{D}\left( \text{DDIM}_{\text{samp}}\left(\text{DDIM}_{\text{inv}}\left(\mathcal{E}(\mathbf{V_0}), T, T^*\right)\right)\right),
\]

Here, \( \mathcal{E} \) signifies the encoding function, \( \mathcal{D} \) the decoding function, and \( \mathbf{V_0} \) the initial video input. The inclusion of both \( T \) and \( T^* \) ensures that the DDIM not only captures the masked patterns but also adheres to the specified activity prompts. This dual guidance mechanism facilitates precise control over the generative process within the latent space, yielding an output video that accurately reflects the desired activities.By operating in this fashion, our model not only retains the fidelity of the original video but also enriches it with detailed descriptions of activities, providing a comprehensive depiction that encompasses both the observed and inferred aspects of the scene.

% Masked training can significantly accelerate the training of large diffusion models like Stable Diffusion without sacrificing generative performance. To accelerate the pace of the learning process, we apply a masked strategy for the designated pattern we want for our sensor description. For example, in figure~\ref{decoder_training} ,  the output fps is 24, we can guide the result to different scenario. Human study is conducted for futherly evaluation

\begin{figure}[h]
  \centering
  \includegraphics[width=1\linewidth,,height=0.2\textheight]{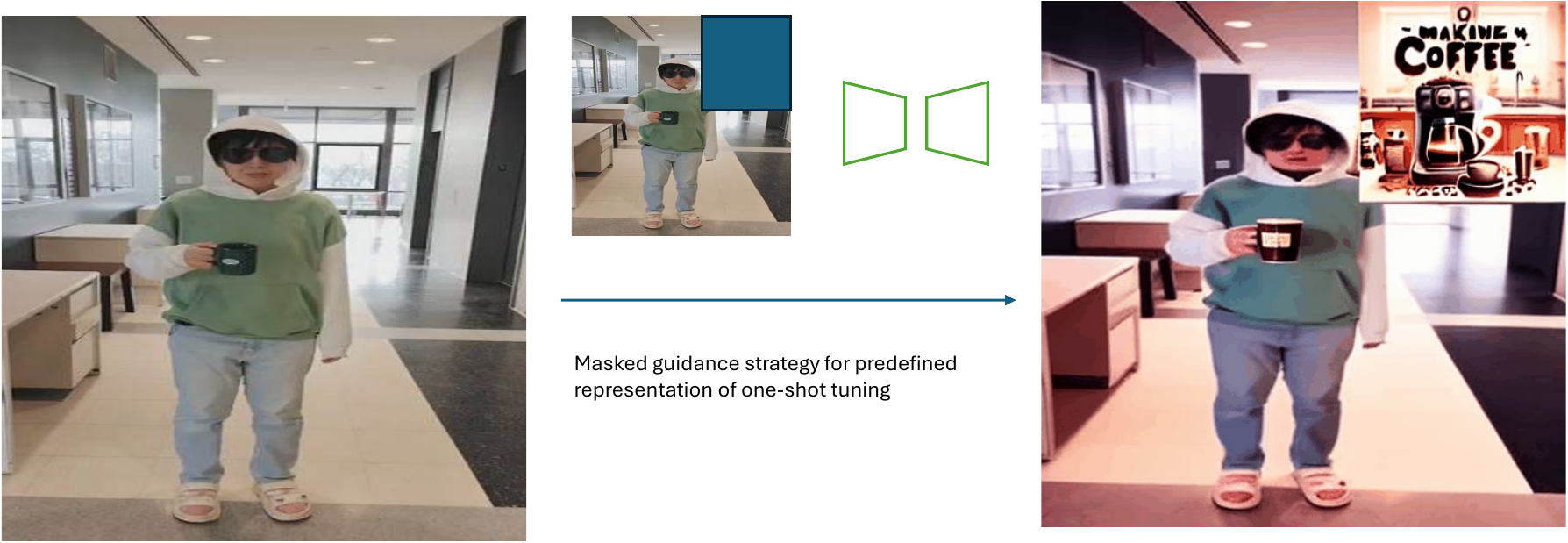}
  \caption{We introduce a mask-guided one-shot tuning strategy designed to seamlessly adapt the model to newly added activities.Example is from opportunity dataset.}

  \label{decoder_training}
\end{figure}

Masked training can markedly expedite the training of substantial diffusion models such as Stable Diffusion\cite{zheng2023fast}, without detriment to their generative capabilities. To hasten the learning process, we employ a masked strategy tailored to specific patterns desired for our sensor descriptions. For instance, as depicted in Figure~\ref{decoder_training}, the output frame rate is set at 24 fps, enabling us to direct the model's output toward varying scenarios. Additionally, a human study is conducted to further evaluate the effectiveness of these strategies.

% \subsubsection{Implementation Details }
% Our decoder is based on  stable diffusion \cite{Rombach_2022_CVPR}, the pretrained weights online can be used, however, there are certain patterns like complex scenario representation pattern especailly needed in smart space scenario is still missing, so we pretrained the absent concept after loading the weights. The pretrain step involves in training all the weights of the model while fine tune  in our case, only tuning several attention layers to implement the  one shot tuning strategy, which lead to  a significant level of gpu memory reduction, can be used on a normal gaming gpu like RTX track gpu card . In our case, we need 10000 steps to pretrain the whole model to learn the new feature while 500 step to tune a video. batch size is 1. The DDIM sampler need 100 steps to inference. we use  NVIDIA A100 GPU for pretrain and use an RTX 6000 for finetuning. we need 48 hours to learn a new feature while 30 minutes for finetune. This reflect advantage for indivdualized  smart space  use case, normal gpu can finish that task of adapting the new feature to the model. This framework makes it implementable in real life case.
 
 \subsubsection{Implementation Details}
Our decoder leverages the stable diffusion model \cite{Rombach_2022_CVPR}, utilizing online pretrained weights. Despite this, specific patterns crucial for representing complex scenarios in smart spaces are absent, necessitating further pretraining of these missing concepts post-weight loading. Our pretraining involves adjusting all model weights, whereas finetuning focuses only on several attention layers to employ a one-shot tuning strategy\cite{wu2023tune}, significantly reducing GPU memory usage. This efficiency allows the use of standard gaming GPUs, like the RTX series. 

Specifically, the complete model requires 10,000 steps for pretraining to incorporate new features, and only 500 steps to fine-tune for video applications, with a batch size of one. For inference, the DDIM sampler requires 100 steps. We utilize an NVIDIA A100 GPU for pretraining and an RTX 6000 for fine-tuning. The entire pretraining process takes approximately 48 hours to integrate a new feature, while finetuning is completed in just 30 minutes. This rapid adaptability is particularly beneficial for individualized smart space applications, where conventional GPUs can effectively handle the adaptation of new features to the model. Such capabilities render our framework practical for real-life implementations.

\section{EXPERIMENTS AND RESULTS}
\label{chap:4}

In this section, we detail the public datasets utilized in our experiments, providing a foundation for a comprehensive evaluation of our methodology. We will systematically discuss the outcomes of each experiment, highlighting the efficacy of our model across different scenarios. Additionally, results from human studies will be presented to demonstrate the practical effectiveness and user perception of our model.  This approach allows us to present a well-rounded assessment of our framework's performance and its applicability in real-world settings.

\subsection{Dataset}  

For our experiments, we utilized three publicly available datasets: \textit{Opportunity}\cite{chavarriaga2013opportunity}, \textit{FallAllD}\cite{saleh2020fallalld}, and \textit{Cooking}\cite{hyzg-9m49-20}. Each dataset offers unique characteristics suited to testing the versatility of our model under different conditions. The \textit{Opportunity}  dataset provide labels for atomic activities corresponding to specific time intervals, facilitating precise activity recognition tests. In contrast, the \textit{Cooking} and the  \textit{FallAid}  datasets, which reflect real-life scenarios, labels only the types of atomic and complex activities within each time interval without specifying the exact timing of each atomic activity. This dataset merely indicates that a group of atomic activities occurs during the specified intervals, presenting a challenge in distinguishing individual actions within these periods.

\subsubsection{Opportunity}
\begin{table}[ht]
\centering

\begin{tabular}{@{}ll@{}}
\toprule
Complex Activities & Atomic Activities \\ \midrule
Making Coffee      & Open Door 1,Close Drawer 1,Open Door 2,Open Drawer 2 \\
Morning Routine & Close Door 1,Close Drawer 2,Close Door 2,Open Drawer 3 \\
Cleaning Up        & Drink,Open Fridge,Close Fridge,Clean Table,Open Dishwasher \\
Making Sandwich  & Close Drawer 3, ,Close Dishwasher,Toggle Switch,Open Drawer 1 \\ \bottomrule
\end{tabular}
\caption{Complex activities and their  atomic activities in the Opportunity dataset. The dataset presents challenges in identifying activities due to the presence of similar atomic activities with different types and the possibility of the same atomic activity belonging to multiple complex activities. This hierarchical and overlapping nature of activities highlights the complexity of activity recognition in real-world scenarios.}
\label{opportunity_dataset}
\end{table}

The Opportunity dataset is a publicly accessible benchmark for human activity recognition algorithms, featuring data from 4 subjects out of an original 12~\cite{vd6r-db31-21} . It includes 15 networked sensor systems with 72 sensors across ten modalities embedded in the environment, objects, and worn on the body. This study particularly focuses on inertial sensors placed on the left lower arm, left upper arm, right lower arm, right upper arm, and back of the torso, which record data via accelerometers, gyroscopes, and magnetometers.

We analyze 17 micro-activities and 4 complex activities that offer a comprehensive view of human motion~(table \ref{opportunity_dataset}). The dataset's design ensures a realistic portrayal of activities, with annotations at multiple abstraction levels suitable for testing advancements in sensor selection, feature extraction, classifier training, multimodal data fusion, segmentation, and hierarchical activity recognition. This rich framework makes the Opportunity dataset an excellent resource for evaluating the effectiveness of different activity recognition strategies under naturalistic conditions.We use 3 subject as training set, 1 subject as testing set.In this study, we utilize a 20-second sliding window approach for data processing, which frames our training set with dimensions of (3036, 72, 600) and our testing set as (1299, 72, 600). This strategy is instrumental in capturing the temporal dynamics essential for our activity recognition tasks.

\subsubsection{FallAllD}

\begin{table}[ht]
\centering
\begin{tabular}{@{}ll@{}}
\toprule
Complex Activities & Atomic Activities \\ \midrule
Falling prediction     & stand, walk, sit, trip \\
Normal ADLs, no Falling detected    & recovery, jog, slip, rotate \\ \bottomrule
\end{tabular}
\caption{Complex activities and their  atomic activities in the Fallaid dataset. The dataset provides information about the types of atomic activities and their  high-level activities. However, it does not include specific labeling for the atomic activities at each time unit. Some atomic activities, such as "walk to the chair and  sit" may belong to both falling prediction and ADL categories, highlighting the hierarchical and overlapping nature of the activities. This poses challenges in accurately labeling and recognizing activities at a granular level.}
\label{fallaid_dataset}
\end{table}

The FallAllD dataset is a specialized resource tailored for research in fall detection, fall prevention, and human activity recognition, suitable for both classical and deep learning methodologies. Data collection involved three identical data-loggers worn by participants on the neck, wrist, and waist, each outfitted with an inertial module (comprising an accelerometer, gyroscope, and magnetometer) and a barometer.

Data were collected from 15 participants aged between 21 to 53 years old, resulting in 26,420 files, each 20 seconds in duration. Within this dataset, complex scenarios such as "fall" and "no fall" detection are highlighted, where atomic activities are indicative of both scenarios. For example, activities might include normal actions such as walking and sitting, as well as the same actions performed with an accidental fall. For our studies, we selected 8 (table ~\ref{fallaid_dataset}) activities that span both complex categories, allocating around 20\% of the data for testing and 80\% for training purposes.In our methodology, we adopted a 10-second sliding window for data segmentation, resulting in a training set dimensionality of (4964, 12, 4760) and a testing set dimensionality of (1375, 12, 4760). This windowing technique is critical for capturing temporal patterns in the sensor data conducive to recognizing complex activities.This dataset does not include specific labels for atomic activities. Each file only provides information on the type of activities, device type, recording date, and subject number.

\subsubsection{Cooking Activity}

\begin{table}[ht]
\centering
\begin{tabular}{@{}ll@{}}
\toprule
Complex Activities & Atomic Activities \\ \midrule
Making Sandwich      & add, cut ,mix \\
Making a Cereal    & open, peel ,pour \\
Making Fruit salad   & put ,take, wash\\ \bottomrule
\end{tabular}
\caption{Complex activities and their associated atomic activities in the Cooking dataset. The dataset represents real-life scenarios and provides information about the types of atomic activities and their corresponding complex activities. However, it does not include specific time labels for each activity. The same atomic activity may belong to different complex activities, highlighting the versatility and reusability of actions across various cooking tasks. This lack of temporal information and the presence of overlapping activities pose challenges in precisely identifying and segmenting individual activities within the dataset.}
\label{cook_dataset}
\end{table}

% The Cooking Activity Recognition Challenge dataset, collected from smartphones, wristwatches, and a motion capture system, presents unique challenges for activity recognition. The data, gathered from 4 subjects, includes accelerometer readings from the right arm, left hip, and both wrists, as well as motion capture data from 29 markers. The dataset encompasses three complex or macro activities (sandwich, fruit salad, and cereal) and ten smaller or micro activities (add, cut, mix, open, etc.).Two main challenges arise from the dataset: varying sampling rates across modalities and missing data, particularly in the left wristwatch accelerometer, Another significant challenge is the random absence of data or missing data, particularly in the accelerometer data from the left wristwatch, which has several samples with no recorded data. The recorded files from the accelerometers and motion capture system have different sampling rates among the 30-second samples or segments.he accelerometers' sampling rates vary between 50 and 100 Hz, while the motion capture samples are captured at approximately 100 Hz. 

The Cooking Activity Recognition Challenge dataset\cite{hyzg-9m49-20} is a multifaceted collection of sensory data, recorded via smartphones, wristwatches, and motion capture systems, designed specifically for the complex task of activity recognition. This dataset, procured from 4 subjects, comprises accelerometer data from the right arm, left hip, and both wrists, in addition to motion capture information from 29 distinct markers. It categorizes activities into three macro activities—making a sandwich, preparing a fruit salad, and cereal preparation—alongside 10 micro activities such as adding, cutting,  mixing and "other". In our recognition task, we focused on nine specific activities that were relevant to our study. We excluded the "other" category from our analysis, as it did not provide meaningful information for our research objectives.The dataset's primary complexities arise from inconsistencies in sampling rates across different sensors and significant instances of missing data, notably within the left wristwatch accelerometer data, which shows several intervals devoid of readings. Furthermore, the dataset's recordings, spanning across the accelerometers and motion capture system, demonstrate varying sampling rates within the 30-second segments. Specifically, accelerometer data sampling rates fluctuate between 50 and 100 Hz, while the motion capture data is consistently sampled at around 100 Hz. These challenges underscore the need for robust processing techniques capable of handling asynchronous data and compensating for informational gaps within the dataset\cite{alia2021summary}.

In our study, we  utilize sensor data recorded from the arm, hip, and wrist, which serve as the primary data inputs for our model. To ensure uniformity across the dataset, we have interpolated all sensor readings to a consistent sampling rate of 100 Hz. Notably, we do not incorporate motion capture data into our analyses.Given the dataset's challenges, including missing readings and inconsistent recording sessions, we have chosen to discretize activities within the same subject. This approach yields a variety of activity sequences, such as "add, cut, open, making cereal" or "cut, take, take, making cereal," each representing different combinations of micro activities that lead up to a macro activity. The dataset presents variability not only in activity sequences but also in sensor availability per session, with some subjects lacking consistent sensor data across recordings. To address these discrepancies and the issue of missing data, we structured the dataset to ensure diversity in the training and testing sets, with distinct activity combinations and previously unseen activity types.It is important to note that the dataset provides only a general indication of atomic activities. Hence, most CHAR models that rely on time-specific atomic activity labels or sequential labeling are not suitable for this dataset. Our approach, tailored for this type of loosely labeled data, allows for robust activity recognition despite the absence of precise temporal annotations.In light of the inseparable nature of the segments within the dataset, we have preserved the original 30-second recording window size to maintain the integrity of the data. Consequently, the shape of our training set is (2326, 12, 3000), indicating that it comprises 2326 segments, each with 12 features, across 3000 time steps. Similarly, the testing set is structured as (711, 12, 3000), consisting of 711 such segments. This configuration ensures that the dataset’s original structure is retained, allowing for an authentic evaluation of our model’s performance.

\subsection{Evaluation}

\subsubsection{ Atomic Accuracy Score}

To assess the precision of our model in detecting atomic activities, we utilize the Atomic Accuracy Score. This metric measures the proportion of atomic activities correctly identified above a specified confidence threshold \( \alpha \) relative to the total number of activities detected. The score is mathematically defined as:

\[
\text{Atomic Accuracy} = \frac{\sum_{i=1}^{n} \chi(p_i > \alpha)}{n}
\]

where \( p_i \) denotes the confidence score of the \( i \)-th detected activity, \( n \) is the total number of detected activities, and \( \chi \) is the indicator function that returns 1 if \( p_i > \alpha \), and 0 otherwise. In practical terms, we set \( \alpha = 0.4 \), ensuring that only activities detected with a confidence level above 40\% are considered in the accuracy measurement. This strategy focuses on the most reliably detected activities, thus providing a more meaningful assessment of the model’s performance in smart space environments.

\subsubsection{ Complex Activity F1 score }

In the evaluation of our model's performance on complex activity classification, the F1 score is employed as a critical metric. The F1 score is a harmonic mean of precision and recall, providing a balanced measure that considers both the false positives and false negatives. This is especially important in our context where some complex activities may be underrepresented in the dataset.

The F1 score is calculated as follows:

\[
\text{F1 Score} = 2 \times \frac{\text{Precision} \times \text{Recall}}{\text{Precision} + \text{Recall}}
\]

\subsubsection{Model Comparison}

In this study, we evaluate the performance of our proposed model against several established baselines in the field of CHAR. We provide a detailed description of our model in Subsection~\ref{sub_section_encoder}. For a fair comparison, we design the baselines with similar structures to ensure that each comparison model has approximately the same number of parameters. We employ the AdamW optimizer for training, with a maximum of 300 epochs. The following subsections detail the configurations of each comparison group:

\begin{itemize}
    \item \textbf{ConvLSTM:} Often considered a foundational architecture for addressing CHAR problems\cite{peng2018aroma,varshney2022human,singh2023acknowledgment,haresamudram2019role}, the ConvLSTM serves as a baseline to evaluate the enhancements our method brings to activity recognition. This model integrates a sensor fusion module utilizing CNN to extract primary sensor features, combined with LSTM networks to capture temporal dependencies among these features. The architecture culminates in a temporal convolution layer followed by a dense output layer, which collectively aim to optimize the recognition of complex human activities.
    \item \textbf{Concept Bottleneck\cite{koh2020concept}:} Utilizes a CNN-based structure to detect atomic concepts before advancing to higher-level concepts, applying MSE loss for concept identification. This hierarchical approach is possible to applied to complex activity recognition.
   \item \textbf{PEMM:} The Pointwise Error Minimization Method (PEMM) is utilized as a comparative baseline in our study. It addresses potential concerns that Concept Bottleneck outcomes are limited by reliance on CNN architectures. To evaluate our model's performance enhancements, we incorporate MSE in an ablation study, contrasting it with PEMM to highlight the effectiveness and advancements of our approach.
    \item \textbf{Debornair\cite{chen2021deep}:} This model parallels our basic ConvLSTM structure with a distinct preprocessing module that processes sensor data for atomic and complex activities separately before merging them. Debornair exclusively design for  complex output only, focusing solely on the integration of processed data without predicting atomic activities.
    \item \textbf{XCHAR\cite{jeyakumar2023x}:} Based on a vanilla ConvLSTM architecture, XCHAR differentiates itself by employing CTC(Connectionist Temporal Classification) loss to emphasize the importance of sequence in atomic activities. However, real-life datasets often suffer from errors or omissions in the sequencing of atomic activities, limiting the applicability of this model to certain datasets where precise sequence labeling is feasible.

\end{itemize}

\begin{table}[htbp]
\centering
\label{tab:model_comparison}
\resizebox{\textwidth}{!}{%
\begin{tabular}{|l|cc|cc|cc|}
\hline
\multirow{2}{*}{Model} & \multicolumn{2}{c|}{Opportunity} & \multicolumn{2}{c|}{Cooking} & \multicolumn{2}{c|}{FallAllD} \\ \cline{2-7} 
 & CHAR F1 Score & Atomic Acc. Score & CHAR F1 Score & Atomic Acc. Score & CHAR F1 Score & Atomic Acc. Score \\ \hline
PEMM & 0.7321 & 0.5332 & 0.7838 & 0.4456 & 0.8217 & 0.7768 \\
ConvLSTM & 0.7922 & -- & 0.8135 & -- & 0.8393 & -- \\
Concept Bottleneck & 0.6479 & 0.4031 & 0.4355 & 0.3728 & 0.5966 & 0.4725 \\
XCHAR & 0.8357 & \textbf{0.6736} & -- & -- & -- & -- \\
DEBONAIR & 0.8015 & -- & 0.8128 & -- & 0.8283 & -- \\
VCHAR & \textbf{0.8463} & 0.6052 & \textbf{0.8198} & \textbf{0.5209} & \textbf{0.8657} & 0.8153 \\ \hline
\end{tabular}
}
% \caption{Comparison of CHAR F1 Score and Atomic Accuracy Score across different models and datasets.Some of ther result is not available because the character of the model can not apply to the dataset. like,require specific time lableling and sequnce labeling, makes it not implementable in the specific real life dataset. }
\caption{Comparative Analysis of CHAR F1 Scores and Atomic Accuracy Across Models and Datasets. Not all results are available due to certain models' incompatibility with datasets that lack specific time or sequence labeling, which is essential for their application in real-world settings.}
\label{encoder_result}
\end{table}

In our study, we utilized a variance-based baseline to assess our model's performance in recognizing complex activities. As shown in Table~\ref{encoder_result}, our method surpasses other models with CHAR F1 Scores of 0.8463 on the Opportunity dataset, 0.8198 on the Cooking Challenge dataset, and 0.8657 on the FallAllD dataset, demonstrating its effectiveness in complex activity recognition. Nevertheless, the model encounters challenges in precisely detecting atomic activities when specific labels are available, recording a slightly lower Atomic Accuracy score of 0.6052, in comparison to XCHAR's 0.6736 on the Opportunity dataset. While our approach mainly focuses in complex activity detection, further investigation into its capabilities for atomic activity recognition remains a point of interest.

Remarkably, our model demonstrates a significant advantage in environments where exact time labeling is absent, such as the Cooking dataset, where our method achieves the highest CHAR F1 Score of 0.8198 and Atomic Accuracy of 0.5209. This suggests that our variance approach adapatally handles datasets lacking detailed temporal annotations better than methods relying on point-wise error minimization like MSE loss, which performs well under conditions of precise labeling but struggles otherwise due to its inherent need for accuracy in individual assessments.The performance on the FallAIID dataset mirrors the trend observed in the Cooking dataset.

Moreover, the comparative performance in the table reveals that some models, including PEMM and Concept Bottleneck, compromise complex activity detection in favor of atomic activity recognition. For instance, while PEMM and Concept Bottleneck are designed to improve atomic detail, they fall short in overall CHAR F1 Scores compared to vanilla ConvLSTM, underscoring a trade-off~\cite{gunning2016broad} that our method avoids. Our model's multi-task recognition capability does not sacrifice the quality of complex activity detection, affirming its balanced approach in simultaneous task management.

\begin{figure}[h]
  \centering
  \begin{minipage}{0.32\linewidth}
    \includegraphics[width=\linewidth,height=0.3\textheight]{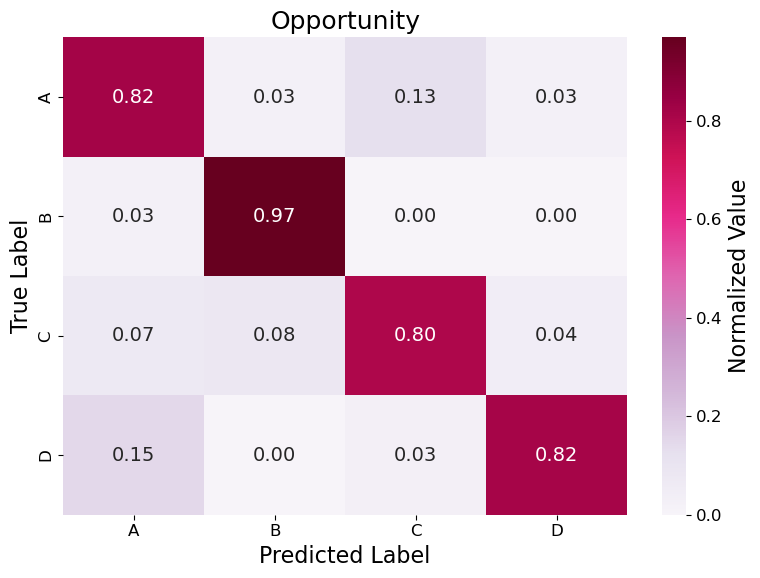}
    \caption{Confusion Matrix of Opportunity dataset, the labels are: A: 'Coffee time', B: 'Early morning', C: 'Cleanup', D: 'Sandwich time'}
    \label{confu_opp}
  \end{minipage}\hfill
  \begin{minipage}{0.32\linewidth}
    \includegraphics[width=\linewidth,height=0.3\textheight]{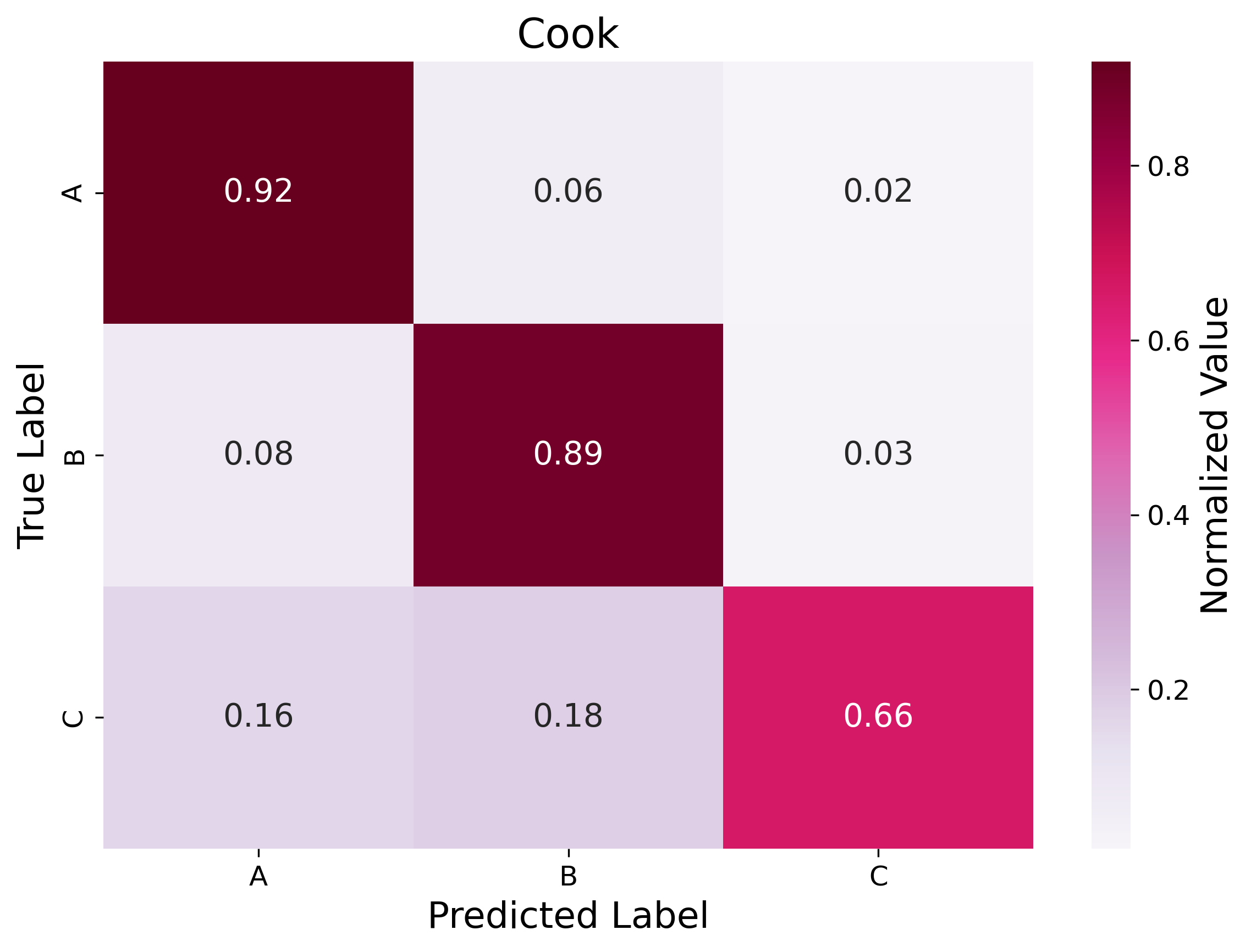}
    \caption{ Cooking Challenge dataset, the labels are: A: 'Making sandwich', B: 'Making fruit salad', C: 'Making cereal'}
    \label{confu_cook}
  \end{minipage}\hfill
  \begin{minipage}{0.32\linewidth}
    \includegraphics[width=\linewidth,height=0.3\textheight]{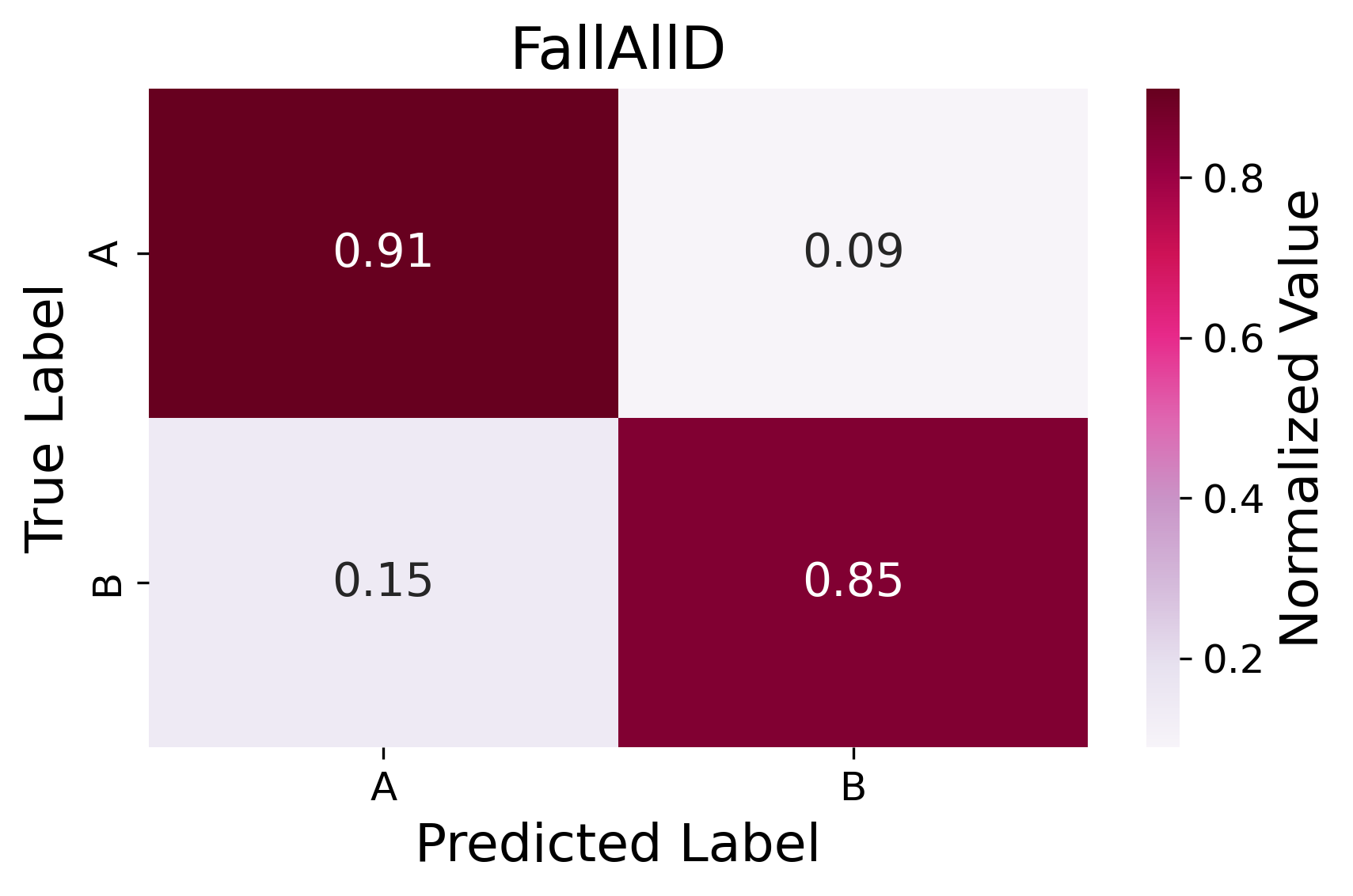}
    \caption{ FallAllD dataset, the labels are: A: 'non-Fall normal ADLs', B: 'Fall detection'}
    \label{confu_fall}
  \end{minipage}
\end{figure}

% \begin{figure}[h]
%   \centering
%   \includegraphics[width=0.3\linewidth,,height=0.3\textheight]{fig/oppo_confusion_matrix.png}
%   \caption{ Confusion Matrix of Opportunity dataset, the labels are: A:'Coffee time', B:'Early morning', C:'Cleanup', D:'Sandwich time' }
%   \label{confu_opp}
% \end{figure}

% \begin{figure}[h]
%   \centering
%   \includegraphics[width=0.3\linewidth,,height=0.3\textheight]{fig/cook_confusion_matrix.png}
%   \caption{ Confusion Matrix of Cooking Challenge dataset,the labels are:  A : 'Making sandwich', B : 'Making fruitsalad', C : 'Making cereal']}
%   \label{confu_cook}
% \end{figure}

% \begin{figure}[h]
%   \centering
%   \includegraphics[width=0.3\linewidth,,height=0.3\textheight]{fig/fallaid_confusion_matrix.png}
%   \caption{Confusion Matrix of FallAllD  dataset,the labels are: A : 'nonFall normal ADLs',  B : 'Fall detection']   }
%   \label{confu_fall}
% \end{figure}

The normalized confusion matrices illustrate distinct performance metrics across datasets. The FallAllD  (fig~\ref{confu_fall}) dataset demonstrates robust detection across all complex activities. For the Opportunity (fig~\ref{confu_opp}) dataset, prediction scores for each complex activity exceed 0.8, with the "early morning" activity achieving the highest accuracy at 0.97, whereas the "clean up" activity records the lowest at 0.8. In contrast, the Cooking Challenge (fig~\ref{confu_cook}) dataset reveals a suboptimal performance with the "making cereal" activity scoring only 0.66, though other activities maintain scores above 0.8.

\subsection{Empirical Studies of Explanation Understandability}

In our effort to develop a user-friendly framework suitable for everyday use by laypersons, we conducted human evaluations to compare our method against existing approaches. These evaluations included all methods tested across the three datasets used in our quantitative experiments. Our evaluation comprised two distinct groups. The first group assessed the clarity and user preference for the model’s output representation, specifically focusing on how effectively the model’s predictions are described and understood by users. The second group of evaluations aimed to demonstrate to users how the model processes different types of data to make decisions. These studies were designed to ascertain which types of explanations regarding the model's decision-making processes are most accessible and favored by users without technical expertise.

In our human study, we utilized three distinct datasets, each evaluated by 100 participants who responded to 6 questions, totaling 1,800 effective responses across all datasets. The participants were recruited predominantly from online platforms, with no specific background prerequisites. Some chose to remain anonymous. Educational backgrounds varied among participants: about one-third had a high school diploma, others held undergraduate or graduate degrees. Ages of participants ranged from twenty to fifty years old.The questions were randomly selected from two categories related to the model's outputs: three questions focused on user preferences for output representation, and three on explanations of model decisions. Participants were asked to rate their satisfaction on a modified Decimal Likert Scale~\cite{wuensch2005likert} ranging from 1 to 5, with options to select intermediate values such as 1.25, 1.5, 1.75, etc., assessing the extent to which the methods enhanced their understanding of the model’s outputs, especially for those without expertise. This enhanced scale provides finer granularity in capturing nuances in participant responses.   This comprehensive evaluation aims to assess various scenarios and input types to better understand user interaction with the model.

\subsubsection{Activity Recognition Description}

We assess the VCHAR, DeXAR, and Concept Bottleneck methods for complex activity recognition, focusing on their ability to represent results effectively within datasets characterized by sparse labeling.

\begin{figure}[h]
  \centering
  \includegraphics[width=0.7\linewidth,]{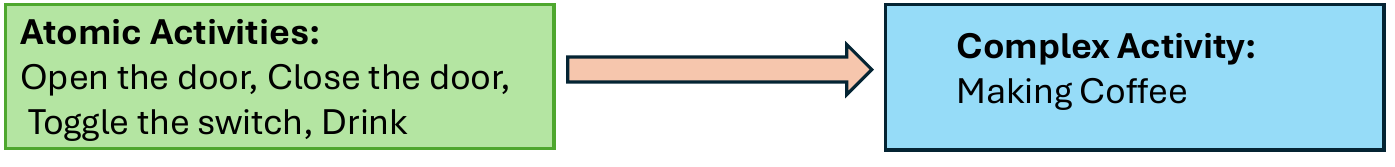}
  \caption{Example of Concept Bottleneck method for vision representation in complex activity recognition.}
  \label{cmb}
\end{figure}

\begin{figure}[h]
  \centering
  \includegraphics[width=0.6\linewidth,]{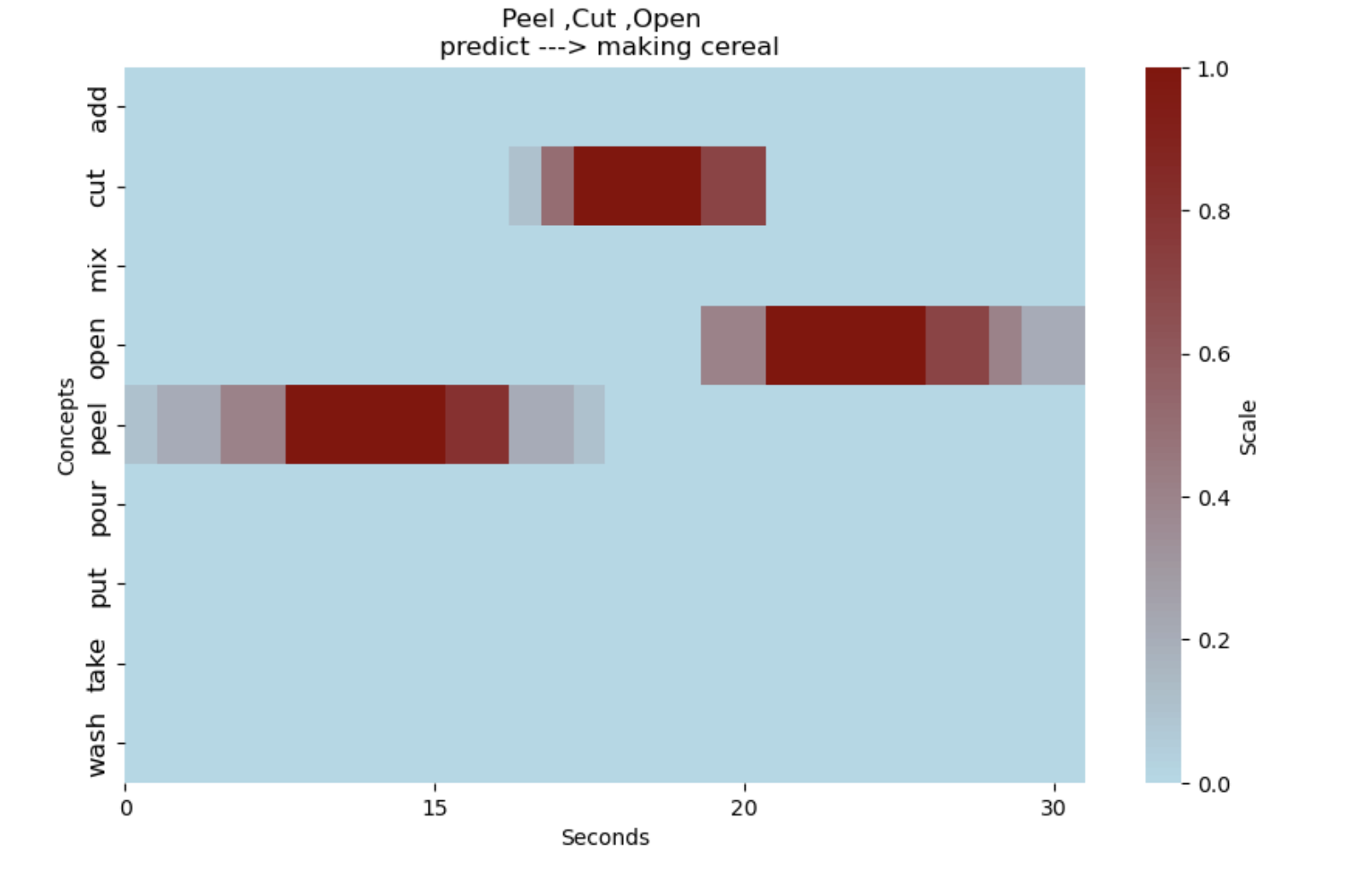}
\caption{DeXAR's semantic image for complex activity detection, highlighting possible atomic activity prediction intervals with a heatmap.}
  \label{dexar}
\end{figure}

\begin{figure}[h]
  \centering
  \includegraphics[width=1\linewidth,,height=0.3\textheight]{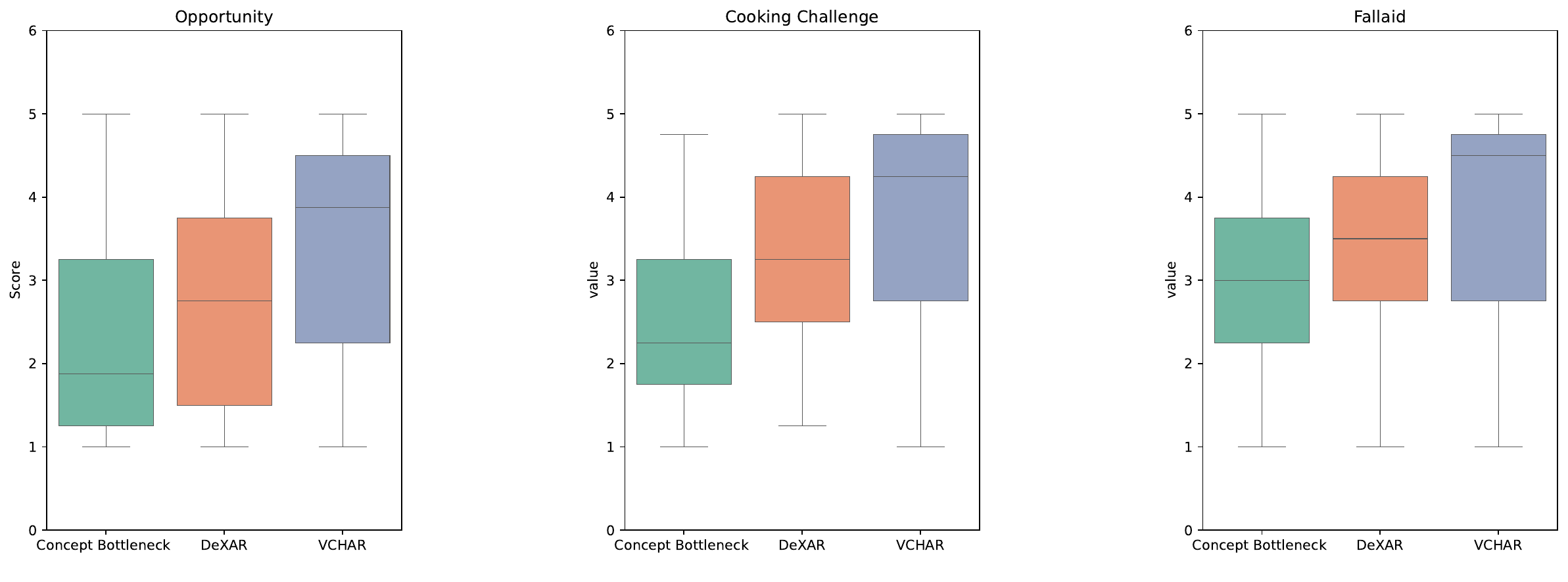}
\caption{Human evaluation of the output description for different model types}

  \label{rep_eval}
\end{figure}

\begin{itemize}
    \item \textbf{DeXAR:}~\cite{arrotta2022dexar} Initially designed for atomic activity recognition, we have modified the method to simultaneously estimate both atomic and complex activities. This adaptation incorporates an NLP-based visual representation to depict the model's recognition results.In our example, as shown in Figure~\ref{dexar}, the semantic visualization employs the Dexar encoding method to represent levels of confidence through color variations. Darker colors indicate higher confidence. This visualization illustrates the potential time intervals estimated by the model for each atomic activity, accompanied by textual explanations. However, it does not provide descriptions for complex activities.

    \item \textbf{Concept Bottleneck:}~\cite{koh2020concept} It merely indicates whether atomic activities are detected or not, without providing detailed information about the recognition results (fig~\ref{cmb}). The output consists only of text descriptions, making it straightforward and concise. Nonetheless, this approach necessitates that users possess basic reading skills.In Figure~\ref{cmb}, we observe that the complex activity is described solely through textual means. While some users may find this approach straightforward and intuitive, it essentially involves a sequence of actions---such as opening and closing a door, toggling a switch, and drinking---that collectively signify a complex activity, in this instance, 'making coffee'.

     \item \textbf{VCHAR:} In Figure~\ref{rep_eval}, the VCHAR system is illustrated, showcasing its capability to represent each atomic activity with a corresponding video and a descriptive label---in this example, 'making cereal'. The width of the video segment indicates the estimated time interval during which the activity occurs, as detected by VCHAR. Both the atomic and complex activities are labeled at the top of the graph. VCHAR is specifically designed to address issues of label sparsity. The time interval estimation is based on the weights connecting a specific last-layer neuron to all neurons in the time series layer, allowing for a comprehensive representation that integrates visual and textual descriptions for each activity.

\end{itemize}

% From figure ~\ref{rep_eval} we can see that , the representation of VCHAR is most preferable. The median evaluation value of VCHAR for each dataset is 3.88,4.37,4.5, we can see from the result that user prefer more descriptions to understand the complex senario, especially for layperson, they focus more on description that apply to daily life, that can provide more descriptions 

As illustrated in Figure~\ref{rep_eval}, VCHAR's representation receives the highest median evaluation scores across the datasets—3.88 for Opportunity, 4.37 for Cooking Challenge, and 4.5 for Fallaid. These scores demonstrate a clear preference among users for more detailed and descriptive outputs to better understand complex scenarios, especially in everyday contexts. The distinct advantage of VCHAR is its ability to convey intricate sensor data interactions in a manner that is intuitive for laypersons. This preference underscores the importance of designing AI systems that not only perform well but also communicate their processes and results in ways that enhance user comprehension and trust in technology applications.

\begin{figure}[h]
  \centering
  \includegraphics[width=0.9\linewidth,]{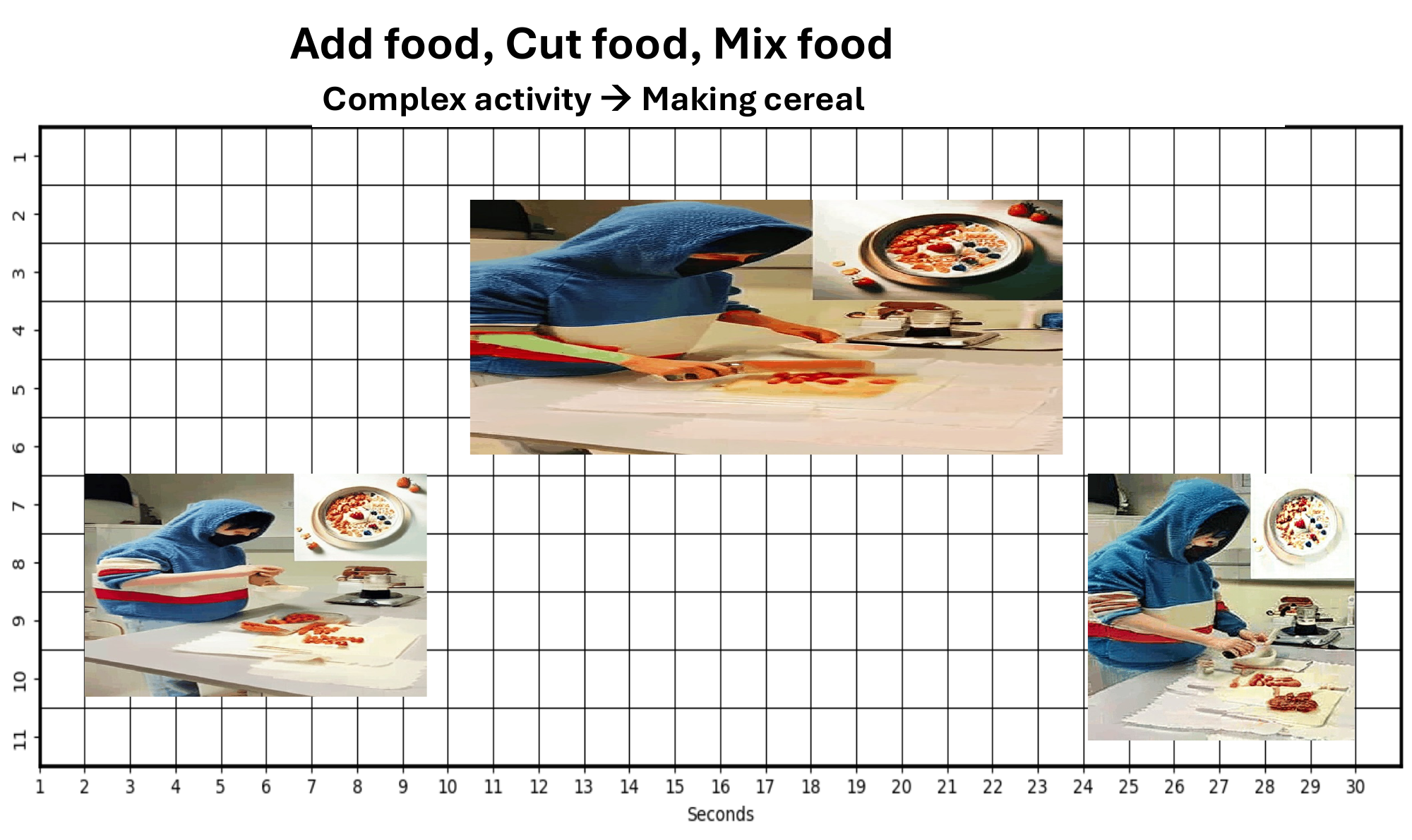}
 \caption{VCHAR depicts atomic and complex activities with videos, using video width to approximate activity prediction timing. The example is from cooking challenge dataset}

  \label{vchar}
\end{figure}

\subsubsection{Model Explaination}

Another aspect of our evaluation focuses on illustrating to users how the model processes various data types to arrive at decisions. We assessed three distinct approaches: our proprietary method, a model-agnostic method, and a model-transparent method. This comparative study seeks to explore how different modeling approaches influence user preferences, particularly in terms of model interpretability and its impact on user satisfaction. Additionally, we analyze various explanation methods to identify which sensors are critical for recognizing atomic activities, further enhancing our understanding of each method's effectiveness in practical scenarios.

\begin{figure}[h]
  \centering
  \includegraphics[width=1\linewidth,height=0.2\textheight]{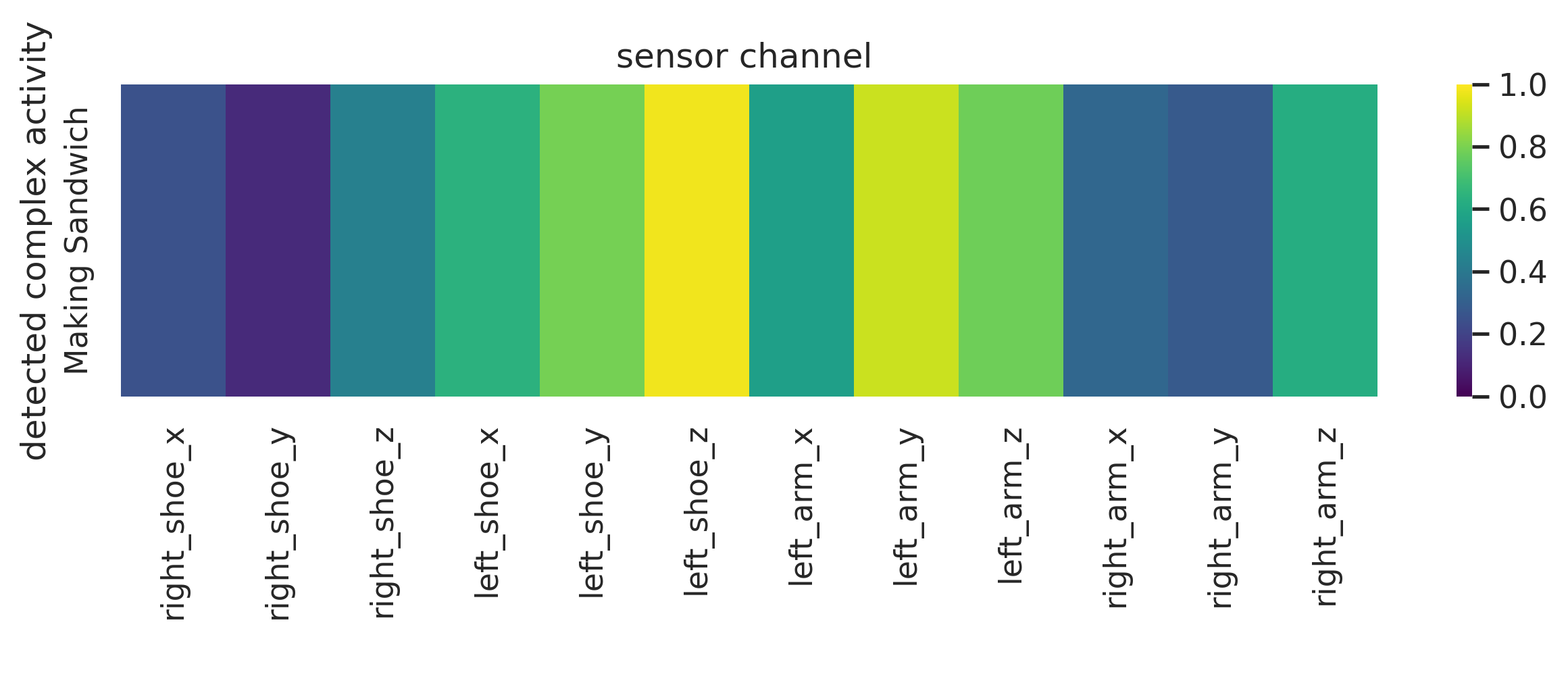}

\caption{Grad-CAM elucidates the decision-making mechanisms within neural networks by mapping the gradients associated with a specific target concept—such as a class output pre-softmax—back to a pertinent hidden layer. Specifically, we employ the deepest hidden layer to assess the contribution of various sensors to the model's predictions.}

  \label{salient_grad}
\end{figure}

\begin{figure}[h]
  \centering
  \includegraphics[width=0.8\linewidth,]{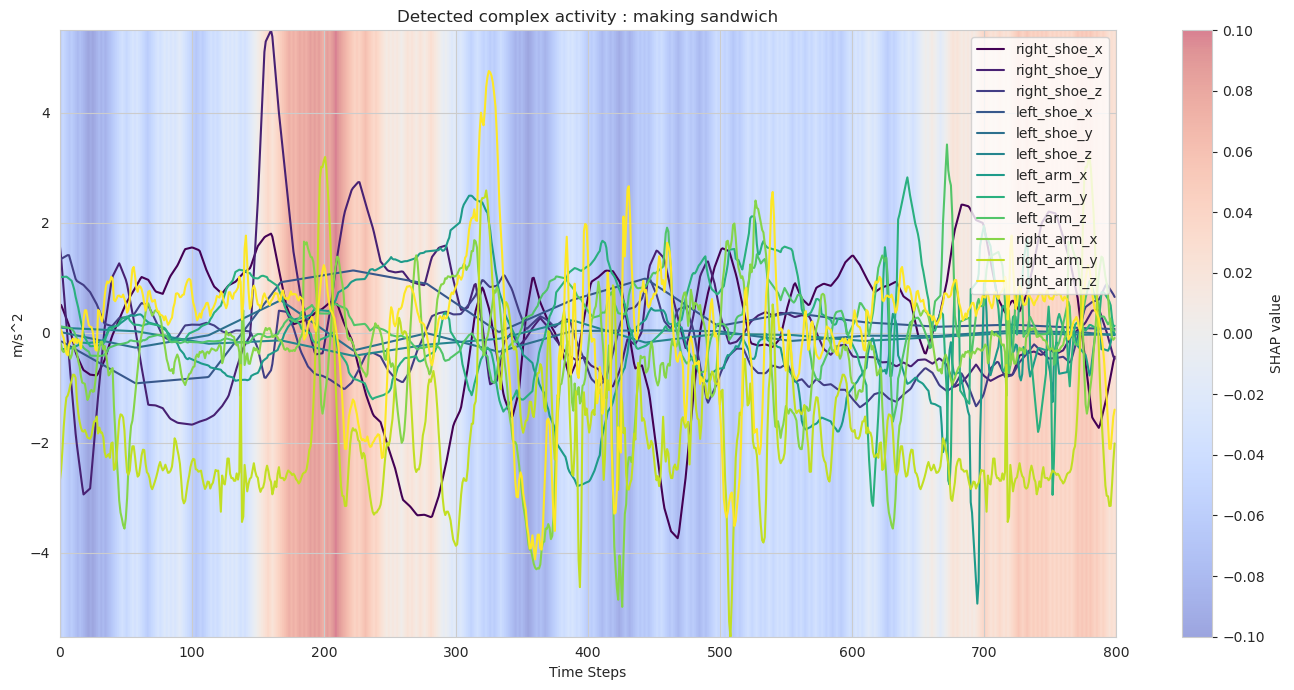}
\caption{The SHAP method primarily approximates the relationship between input signals and output predictions without probing the model's internal mechanisms. Explanations are directly applied to the input data.}

  \label{shap}
\end{figure}

\begin{figure}[h]
  \centering
  \includegraphics[width=0.5\linewidth,height=0.4\textheight]{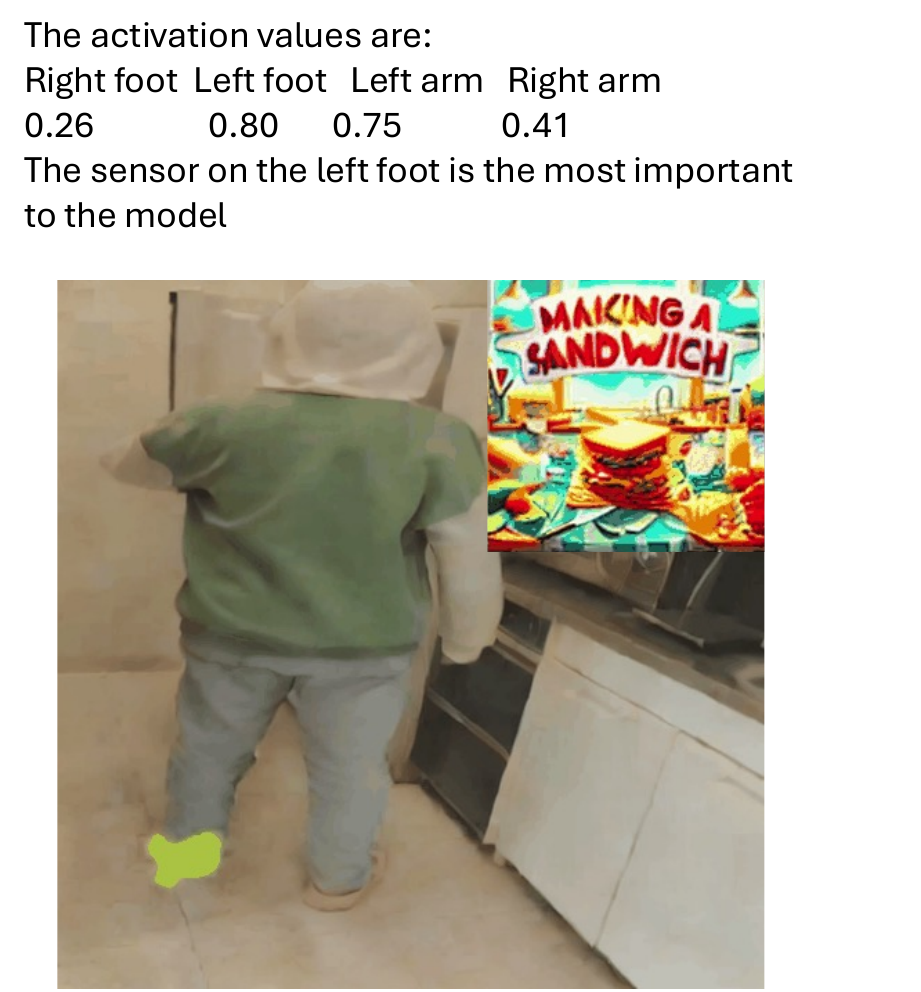}

% \caption{The image on the left exclusively visualizes the atomic activity, while the image on the right concurrently visualizes both the atomic activity "cutting the food" and the complex activity "making cereal." In our approach, the decoder is trained to generate specific colors to highlight the corresponding sensor locations that are most crucial to the model's decision-making process. The values represent the average measurements across the channels of a single sensor.}

\caption{The figure illustrates atomic activity associated with "opening a refrigerator", as part of the complex activity "making a sandwich". The fluorescent yellow color on the left foot indicates significant sensor contributions from this location when the model detects the activity. Our method trains a decoder to assign specific colors to highlight critical sensor positions crucial for the model’s decision-making. The depicted values are average readings from the channels of a single sensor.}

  \label{vhar_act}
\end{figure}

\begin{itemize}
\item \textbf{Grad-CAM:}~\cite{lundberg2017unified} Grad-CAM is a model-transparent method that calculates the gradient of a target concept (output) relative to the feature maps of a designated  layer. It produces a heatmap that identifies the critical sensors in higher layers that are pivotal for class prediction. As illustrated in Figure~\ref{salient_grad}, it shows how various sensors contribute at different intensities to a particular activity, with brighter areas indicating greater importance.

\item \textbf{SHAP}~\cite{chattopadhay2018grad} In contrast, SHAP is a model-agnostic method that approximates the relationship between inputs and outputs without probing the model’s internal mechanisms. It focuses on representing the sensor signals in the input time series data. SHAP calculates scores by estimating the impact of each feature on the prediction, using Shapley values from cooperative game theory to quantify each feature's contribution.Shapley values in SHAP can be positive or negative, depending on whether a feature contributes positively or negatively to the model's prediction for a particular instance, relative to the average prediction. This method highlights critical regions relative to the predicted class, as illustrated in Figure~\ref{shap}. Given the inherent complexity of sensor signals, simple descriptions of input may not fully capture their significance.

\item \textbf{VCHAR} VCHAR delivers a detailed depiction of sensor contributions by integrating both textual and visual representations. As shown in Figure~\ref{vhar_act}, it visualizes a scenario where a person is attempting to open a fridge, with the most significant sensor activation value on the left foot, which VCHAR highlights in the visualization. Additionally, the sensor values are derived from gradient values tied to the selected activity type, akin to Grad-CAM's methodology but extended to include both atomic and complex activities. VCHAR not only identifies these activities but also labels them in the visual representation, enhancing its analytical capabilities for detailed activity analysis.

\end{itemize}

\begin{figure}[h]
  \centering
  \includegraphics[width=1\linewidth,,height=0.3\textheight]{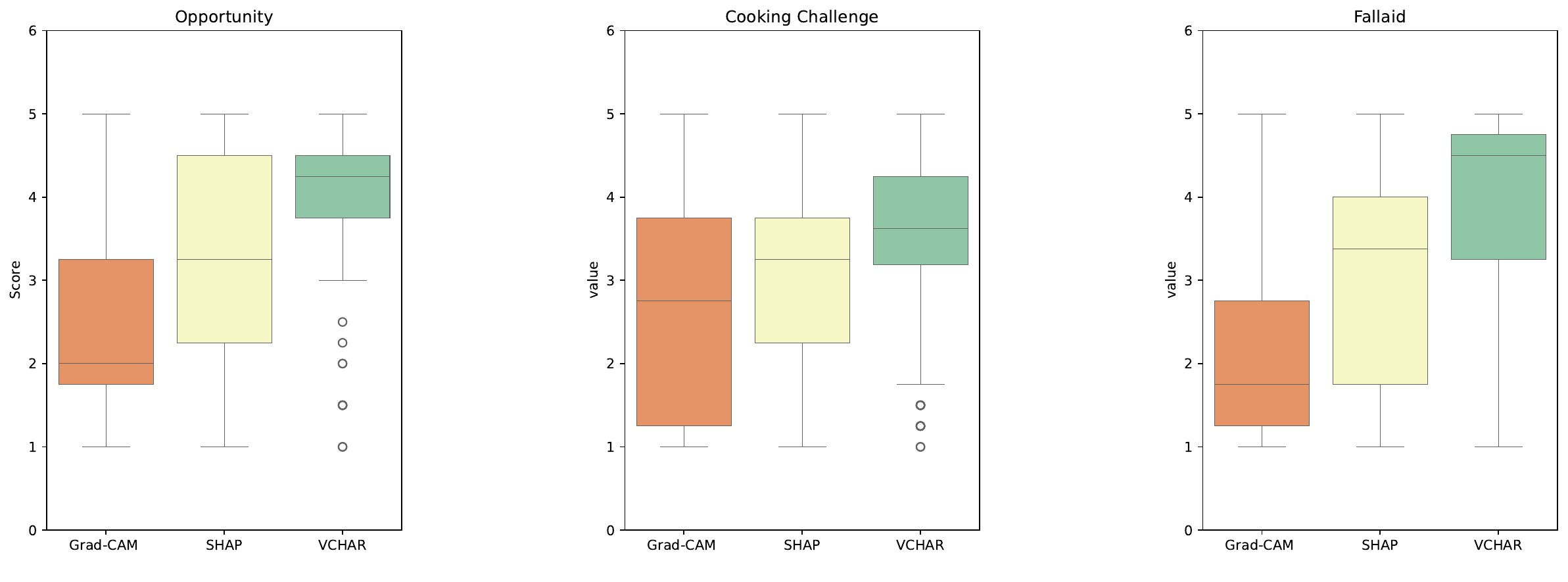}
\caption{Human Evaluation of Sensor Contributions Across Different Methods}

  \label{act_eval}
\end{figure}

From the results depicted in Figure~\ref{act_eval}, VCHAR demonstrates superior performance with median scores of 4.2, 3.8, and 4.5, along with a lower variance. Interestingly, our results are comparable to those of Grad-CAM. Grad-CAM scored lower than SHAP, which primarily stems from SHAP's approach of explaining outcomes based on time-series data, thus providing a more detailed informational context. Despite this, our method, which also employs a model-transparent approach akin to Grad-CAM, received higher scores. This suggests that laypersons may benefit from more detailed explanations provided by our model, as compared to those used by expert users.

\section{\textsc{Conclusions}}
\label{chap:6}

In this paper, we introduce VCHAR, a variance-based method specifically designed to address label sparsity issues in in-the-wild datasets. VCHAR is capable of simultaneously detecting both complex and atomic activities, without compromising the recognition rate of complex activities. Our results demonstrate a performance improvement over other baseline methods. Additionally, we present a novel decoder that translates the model's outputs into a visual representation. This enhancement significantly aids laypersons in understanding the mechanisms of the model, compared to methods traditionally tailored for experts. A human study confirms that our method is preferred over others, offering more accessible and detailed insights to layperson users.

\subsubsection{Limitations }

Although we have made thorough attempts, our study presents several constraints as outlined below:

\begin{itemize}

\item \textbf{Atomic Activity}: This paper introduces a method designed to address the challenges of accurately labeling the timing and sequence of activities in wild datasets, where precise annotation is often lacking. While this approach enhances labeling accuracy, it does not yet achieve optimal accuracy rates compared to precise labeling method.

\item \textbf{Real-Time Rendering}: Our approach features a generative decoder that visually represents activities detected within smart spaces. However, despite its innovative design, the rendering time is extended relative to traditional end-to-end decoding methods. This delay is largely due to the stable diffusion process employed by the decoder, which requires 100 steps to complete the inference. While this method provides detailed visual outputs, it still needs improvement for real-time applications.

\item \textbf{Cross Domain Encoder:} This paper introduces an approach using a universally pretrained decoder to interpret different scenarios within a single model, enhancing the ability to translate various types of scenarios seamlessly. However, while the decoder supports multi-scenario translation, the encoder processes different types of data separately and struggles to recognize cross-domain sensor data as a unified model, limiting its effectiveness in integrated scenario analysis.

\end{itemize}

\subsubsection{Future Work}

% In future research, even our main purpose is to detect complex activities. Enhancing the atomic activity detection rate will still be  an useful research to improve the atomic activity detection rate.
% our primary goal will be to refine and enhance the methods used to identify atomic activities, targeting a significant improvement in accuracy rates. Overcoming the limitations noted in real-time rendering remains a critical challenge; achieving faster rendering times is essential for practical real-world applications. As part of these efforts, we plan to develop more efficient algorithms that can handle the computational demands of stable diffusion processes to generate smart space sensor representation without compromising on the quality and speed of the output.

In future studies, while our principal focus remains on detecting complex activities, improving the detection rates of atomic activities will also be a key area of research. Our primary objective will be to refine the methods used for identifying atomic activities, aiming for substantial enhancements in accuracy. Additionally, addressing the challenges in real-time rendering is crucial; reducing rendering times is imperative for the practical deployment of our models in real-world settings. To achieve this, we plan to develop more efficient algorithms capable of managing the computational demands of stable diffusion processes. These improvements will aim to optimize the generation of smart space sensor representations, ensuring high-quality outputs without sacrificing speed or efficiency.

Additionally, to address the complexities of applying these techniques in diverse real-life environments, we aim to design and implement a unified encoder. This advanced encoder will be capable of processing and translating various types of sensor data across multiple domains into a coherent visual output. The development of such a encoder will facilitate a more seamless integration of our methods into everyday technology, making smart space technologies more adaptable and user-friendly across different settings and applications.

\FloatBarrier

\bibliographystyle{ACM-Reference-Format}
\bibliography{sample-base}

\end{document}